\documentclass{article}

\usepackage{styles/jfrExamplee}
\usepackage{graphicx}
\usepackage{subfig}
\usepackage{setspace}
\usepackage{epstopdf}
\usepackage{enumitem} 		
\usepackage[tight]{units} 	
\usepackage{amsmath}		
\usepackage{amssymb}		
\usepackage{algorithm2e}	
\usepackage{gensymb}		
\usepackage{booktabs}  		
\usepackage{multirow}		
\usepackage{caption}		
\DeclareMathAlphabet{\pazocal}{OMS}{zplm}{m}{n}
\DeclareMathAlphabet{\mathcal}{OMS}{cmsy}{m}{n} 
\usepackage[toc,page]{appendix}		
\usepackage[bottom]{footmisc}		
\hypersetup{
    colorlinks = true,
    citecolor=blue
}
\usepackage{mathtools}		
\usepackage{bm}				
\usepackage[capitalize,noabbrev]{cleveref} 
\usepackage[nolist,nohyperlinks]{acronym}
\usepackage{array}
\usepackage{pseudocode}  	

\crefname{equation}{Eq.}{Eqs.}
\Crefname{equation}{Equation}{Equations}
\crefname{appsec}{Appendix}{Appendices}
\crefname{pseudocode}{algorithm}{algorithms}
\Crefname{pseudocode}{Algorithm}{Algorithms}

\usepackage{color} 
\usepackage{soul} 
\sethlcolor{green}
\definecolor{OliveGreen}{rgb}{0,0.6,0}

\DeclareMathAlphabet{\pazocal}{OMS}{zplm}{m}{n}
\let\oldnl\nl
\newcommand{\nonl}{\renewcommand{\nl}{\let\nl\oldnl}}

\def\code#1{\texttt{#1}}

\makeatletter
\newcommand{\nosemic}{\renewcommand{\@endalgocfline}{\relax}}
\newcommand{\dosemic}{\renewcommand{\@endalgocfline}{\algocf@endline}}
\renewcommand\paragraph{\@startsection{paragraph}{4}{\z@}{-1.5ex plus -0.5ex minus -.2ex}{0.5ex plus .2ex}{\normalsize\itshape\raggedright}}
\makeatother

\setlist[itemize]{itemsep=0pt,parsep=3pt,partopsep=0pt, topsep=3pt, leftmargin=25pt,rightmargin=25pt}
\setlist[enumerate]{itemsep=0pt,parsep=3pt,partopsep=0pt,topsep=0pt,leftmargin=25pt,rightmargin=25pt}

\newcolumntype{L}[1]{>{\raggedright\arraybackslash}p{#1}} 

\title{Meteorology-Aware Multi-Goal Path Planning for Large-Scale Inspection Missions with Long-Endurance Solar-Powered Aircraft}

\author{
\begin{minipage}{35em}
\begin{center}
Philipp Oettershagen, Julian F{\"o}rster, Lukas Wirth, Jacques Amb{\"u}hl and Roland Siegwart\\
\end{center}
\end{minipage} \\
\\
Autonomous Systems Lab\\Swiss Federal Institute of Technology Zurich (ETH Zurich)\\Leonhardstrasse 21\\ 8092 Zurich\\+41 44 632 7395\\
\texttt{philipp.oettershagen@mavt.ethz.ch} \\
}

\begin{document}

\maketitle

\begin{abstract}
Solar-powered aircraft promise significantly increased flight endurance over conventional aircraft. While this makes them promising candidates for large-scale aerial inspection missions, their structural fragility necessitates that adverse weather is avoided using appropriate path planning methods. This paper therefore presents MetPASS, the Meteorology-aware Path Planning and Analysis Software for Solar-powered UAVs. MetPASS is the first path planning framework in the literature that considers \emph{all} aspects that influence the safety or performance of solar-powered flight: It avoids environmental risks (thunderstorms, rain, wind, wind gusts and humidity) and exploits advantageous regions (high sun radiation or tailwind). It also avoids system risks such as low battery state of charge and returns safe paths through cluttered terrain. MetPASS imports weather data from global meteorological models, propagates the aircraft state through an energetic system model, and then combines both into a cost function. A combination of dynamic programming techniques and an A*-search-algorithm with a custom heuristic is leveraged to plan globally optimal paths in station-keeping, point-to-point or multi-goal aerial inspection missions with coverage guarantees. A full software implementation including a GUI is provided. The planning methods are verified using three missions of ETH Zurich's \emph{AtlantikSolar} UAV: An 81-hour continuous solar-powered station-keeping flight, a \unit[4000]{km} Atlantic crossing from Newfoundland to Portugal, and two multi-glacier aerial inspection missions above the Arctic Ocean performed near Greenland in summer 2017. It is shown that integrating meteorological data has significant advantages and is indispensable for the reliable execution of large-scale solar-powered aircraft missions. For example, the correct selection of launch date and flight path across the Atlantic Ocean decreases the required flight time from 106 hours to only 52 hours.
\end{abstract}

\begin{acronym}
\acro{EKF}{Extended Kalman Filter}
\acro{UKF}{Unscented Kalman Filter}
\acro{SLAM}{Simultaneous Localization And Mapping}
\acro{IMU}{Inertial Measurement Unit}
\acro{2D}{two-dimensional}
\acro{2.5D}{2.5-dimensional}
\acro{3D}{three-dimensional}
\acro{6D}{six-dimensional}
\acro{LiDAR}{Light Detection And Ranging}
\acro{UAV}{Unmanned Aerial Vehicle}
\acro{STC}{Standard Test Conditions}
\acro{ESC}{Electronic Speed Control}
\acro{MPPT}{Maximum Power Point Tracker}
\acro{AoI}{Angle of Incidence}
\acro{FM}{Full Model}
\acro{CAM}{Conceptual Analysis Model}
\acro{CDM}{Conceptual Design Model}
\acro{CFD}{Computation Fluid Dynamics}
\acro{DEM}{Digital Elevation Model}
\acro{DP}{Dynamic Programming}
\acro{DoF}{Degrees of Freedom}
\acro{ABL}{Atmospheric Boundary Layer}
\acro{FEM}{Finite Element Method}
\acro{PDE}{Partial Differential Equation}
\acro{NWP}{Numerical Weather Prediction}
\acro{ROS}{Robot Operating System}
\acro{RMSE}{Root Mean Square Error}
\acro{COSMO}{Consortium for Small-Scale Modeling}
\acro{ECMWF}{European Centre for Medium-Range Weather Forecasts}
\acro{ZHAW}{Consortium for Small-Scale Modeling}
\acro{TSP}{Traveling Salesman Problem}
\acro{TDATSP}{Time Dependent Asymmetric Traveling Salesman Problem}
\acro{SPP}{Shortest Path Problem}
\acro{ZHAW}{Consortium for Small-Scale Modeling}
\acro{SoC}{State of Charge}
\acro{GPS}{Global Positioning System}
\acro{OMPL}{Open Motion Planning Library}
\acro{HIL}{Hardware-in-the-loop}
\acro{SAR}{Search and Rescue}
\acro{PID}{Proportional-Integral-Derivative}
\acro{FCL}{Flexible Collision Library}
\end{acronym}

\section{Introduction}
\label{sec:Introduction}
%
\paragraph{Motivation}
\label{sec:Intro_Motivation}

\begin{figure}[htb]
    \centering
    \includegraphics[width=1.0\linewidth]{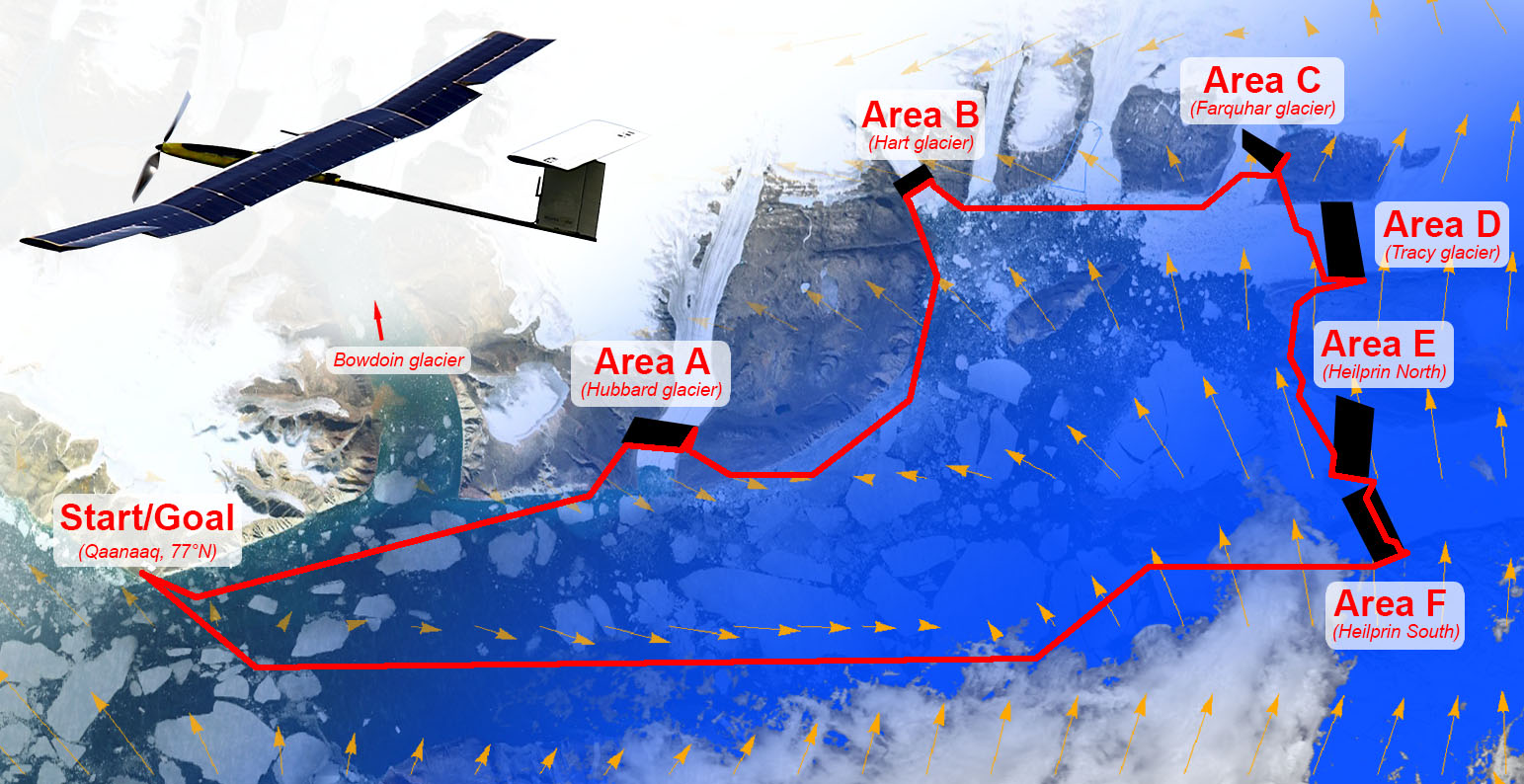}
    \caption{
    Large-scale aerial sensing missions with fragile solar-powered aircraft such as \emph{AtlantikSolar} (top left) require careful route planning. MetPASS, the Meteorological Path Planning and Analysis Software for Solar-powered UAVs presented herein, allows to plan optimal point-to-point routes as well as multi-goal routes that allow inspecting multiple areas of interest --- for example a set of Arctic glaciers --- in one flight. Routes optimized with MetPASS are safe and efficient given the consideration of terrain data and meteorological risk factors such as wind, rain, thunderstorms and clouds.}
    \label{fig:Intro_Collage}
\end{figure}

Solar-powered Unmanned Aerial Vehicles (UAVs) promise significantly increased flight endurance over conventional aircraft. This greatly benefits applications such as large-scale disaster relief, border patrol or aerial inspection in remote areas~\cite{NASA_Pathfinder}. Research and development of solar UAVs is ongoing in both academia \cite{Technion_SunSailor,Rojas_GreenFalconSolarUAV} and industry \cite{AeroVironment_SolarPoweredPuma, SilentFalcon_SilentFalcon}. Since 2005, multiple solar aircraft have been able to demonstrate multi-day continuous flight \cite{Cocconi_SoLong, Noth_PhD,SolarImpulse_5DayFlight}, with the current world record being a 14-day continuous flight~\cite{QinetiQ_Zephyr14dayRecord}. In our own work~\cite{Oettershagen_JFR2017}, we have previously shown an 81-hour continuous solar-powered flight with the \emph{AtlantikSolar} UAV (\cref{fig:Intro_Collage}) which set the current world record in flight endurance for aircraft below \unit[50]{kg} total mass.

However, all solar-powered aircraft necessarily require suitable weather conditions for such long-endurance operations. Thunderstorms, rain and wind gusts can quickly become elementary threats to the aircraft's integrity. Moreover, clouds and strong winds can significantly reduce the solar power income or increase the required propulsion power such that a landing is required. Solar aircraft, or more generally speaking all aerial vehicles that are sensitive to weather e.g. because they are flying slowly or are structurally fragile, therefore require careful pre-operational planning. Safe and efficient flight requires the consideration of terrain, the internal system state (e.g. battery state of charge), and major weather phenomena (thunderstorms, rain, winds and wind gusts, radiation and clouds). A model-based path planning framework (\cref{fig:Intro_Collage}) allows to integrate these effects in a structured manner and is thereby able to generate globally optimal paths for the aerial vehicle.

\paragraph{State of the art}
\label{sec:Intro_StateOfTheArt}

Literature on path planning for solar-powered aircraft is sparse. The necessary elements (e.g. wind-dependence, sun-dependence) are only considered separately in relatively unrelated fields of research. On one hand, when focussing only on wind-aware planning a comprehensive body of research is available. For example, Rubio~\cite{Rubio_2003} proposes evolutionary approaches and considers forecasted 3D wind fields in a large-scale Pacific crossing mission. A planner for oceanic search missions, which employs meteorological data such as temperature and humidity to predict icing conditions, is presented in subsequent work~\cite{Rubio_2004}. Chakrabarty \cite{Chakrabarty2013PlanningDynWindFields} presents a sampling-based planning method for paths through complex time-varying 3D wind fields. All these contributions focus on either fuel-powered or non-solar gliding aircraft. On the other hand, research which does incorporate solar models unfortunately often neglects wind and thus solves a very simplified problem. Rather theoretical approaches are proposed: For example, Klesh \cite{Klesh_2007, Klesh_2009} uses optimal control techniques to generate paths maximizing the UAV's final energy state in small-scale point-to-point and loitering problems. The sun position is however assumed constant. Spangelo~\cite{spangelo2009periodic} investigates optimal climb and descent maneuvers during loitering. Hosseini \cite{Hosseini_2013} adds sun-position time dependence, but assumes perfect clear-sky conditions instead of considering meteorological data. Dai~\cite{Dai_2013} avoids the clear-sky assumption by deriving an expected solar radiation income based on local precipitation and humidity forecasts. The path planning problem is then solved by a Bellman-Ford algorithm. Overall, there is no literature that covers all the system-related or meteorological aspects that can, as described before, affect solar-powered aircraft. In addition, there is a clear lack of flight test based verification of the developed planning approaches.

\paragraph{Contributions}
\label{sec:Intro_Contributions}


This paper presents the Meteorology-aware Path Planning and Analysis Software for Solar-powered UAVs (MetPASS), the first path planning framework in the literature that considers all safety and performance relevant aspects (terrain, system state, meteorological environment) of solar-powered flight. It is able to optimize large-scale station-keeping, point-to-point and multi-goal missions (\cref{fig:Intro_Collage}) and focuses on real-world applications. More specifically, this paper and the corresponding framework present and implement the following contributions:

\begin{itemize}
\item an \emph{optimization approach} that yields cost-optimal aircraft paths by combining an extended A*-algorithm for multi-goal order optimization, a dynamic programming based point-to-point planner and local scan path planner that guarantees area coverage based on a simple camera model. 
\item a \emph{cost function} for solar-powered aircraft. Both safety and performance are considered: The cost function assesses terrain collision risk, the system state (time since launch, battery state of charge, power consumption and generation) through a comprehensive energetic model, and up-to-date meteorological data (thunderstorms, precipitation, humidity, 2D winds, gusts, sun radiation and clouds) through global weather models.
\item a \emph{full software implementation} that features ease-of-use through a GUI and is optimized for computational speed via a custom-designed heuristic and the use of parallelization and caching. As a result, the framework can be used for detailed mission feasibility analysis, pre-flight planning, and in-flight re-planning once updated weather data is available.
\item an extensive \emph{flight-test based verification} of the planning results for the 81-hour flight endurance world record of \emph{AtlantikSolar}~\cite{Oettershagen_JFR2017}, a crossing of the Atlantic Ocean from Newfoundland to Portugal, and two multi-goal glacier inspection missions above the Arctic Ocean near Greenland.
\end{itemize}

The point-to-point path planning approach was already presented in our previous work~\cite{Wirth_AeroConf2015}. The novel aspects extended in this paper are the terrain avoidance, the local scan path planning that guarantees coverage using a camera model, the multi-goal path optimization approach and the flight test results.


The remainder of this paper is organized as follows: \Cref{sec:Design} presents the fundamentals of the optimization approach, i.e. the cost function, heuristic, the dynamic programming based point-to-point path planning and the A*-based multi-goal optimization. \Cref{sec:Implementation} describes the software implementation and verification. \Cref{sec:Results} presents the planning results, compares them to the aforementioned flight data, and provides a computation time analysis. \Cref{sec:Conclusion} provides concluding remarks.

\section{Design}
\label{sec:Design}

\begin{figure}[htb]
    \centering
    \includegraphics[width=1.0\linewidth]{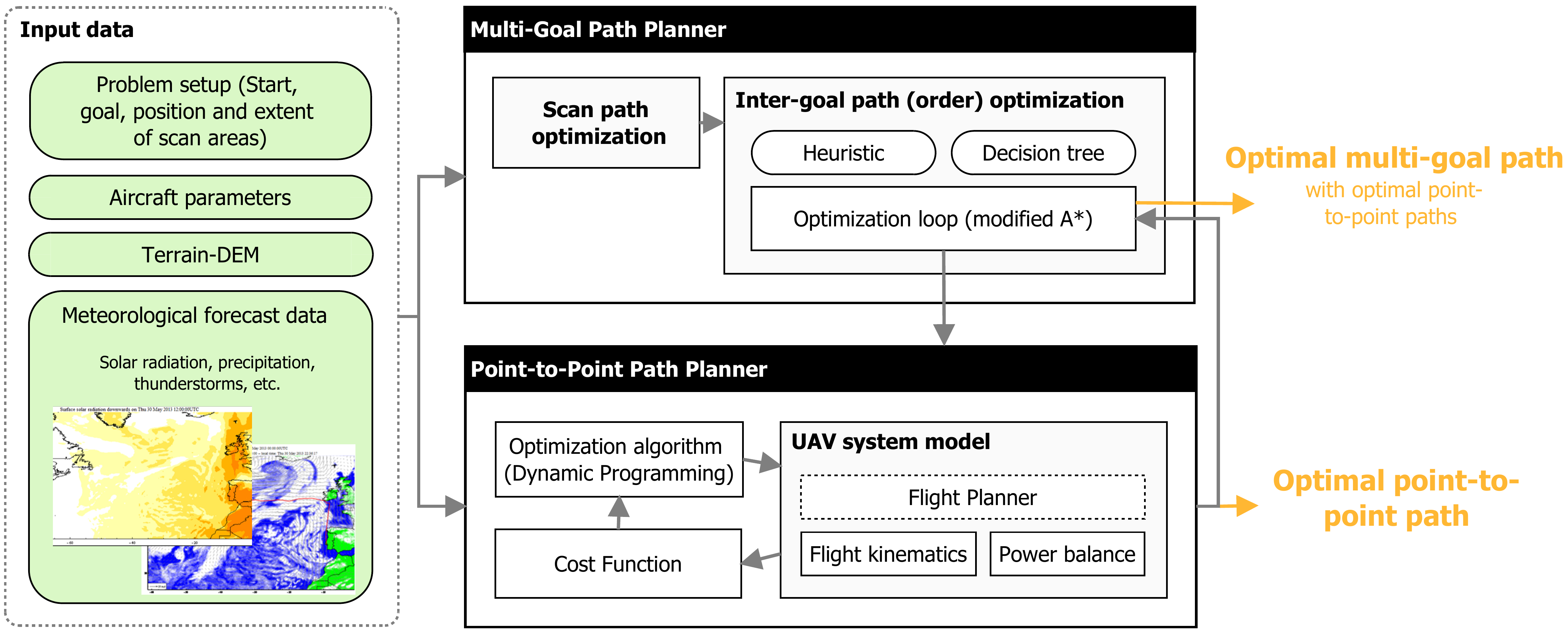}
    \caption{
    MetPASS architecture. Using input data such as meteorological weather forecasts, MetPASS can calculate optimal point-to-point as well as multi-goal paths that allow the inspection of multiple areas of interest.}
    \label{fig:Design_OverviewFlowchart}
\end{figure}

MetPASS plans safe and efficient large-scale aerial inspection missions for fixed-wing aircraft and solar-powered UAVs in particular. Such missions can either be multi-goal missions which inspect multiple areas of interest (\cref{fig:Intro_Collage}) or point-to-point missions. The MetPASS architecture is shown in \cref{fig:Design_OverviewFlowchart}. The \emph{input data} is loaded first. Then, in the multi-goal mission case, the \emph{multi-goal path planner} performs the \emph{scan path optimization} to determine the exact scan path within each area of interest. Using these scan paths, the \emph{inter-goal path optimization} then determines the order in which the areas of interest shall be visited. This process internally calls the \emph{point-to-point planner} to optimize the individual routes between areas of interest.

\subsection{Environment-aware Point-to-Point Path Planning}
\label{sec:Design_P2P}


The optimization problem solved by the MetPASS point-to-point path planner (\cref{fig:Design_OverviewFlowchart}) can be stated as follows: Given fixed departure and arrival coordinates, find a path which minimizes the total cost as defined by a cost function. The departure time can be either a fixed or free parameter. The cost function includes a term for the proximity to terrain, environmental conditions such as solar radiation or precipitation, and aircraft states like power consumption or \ac{SoC}. Environmental conditions are estimated based on time-varying meteorological forecast data in a three-dimensional grid. The system model includes flight kinematics with respect to horizontal wind, power generation through the solar modules and the power consumption of the aircraft. 

\subsubsection{Optimization Algorithm}
\begin{figure}[htb]
    \centering
    \includegraphics[width=0.8\linewidth]{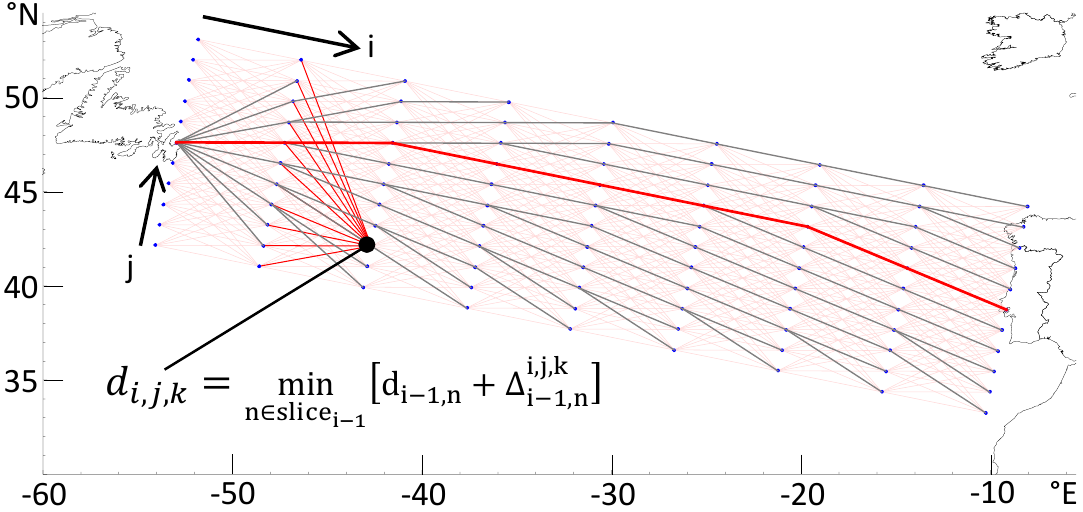}
    \caption{
    Exemplary route optimization from start (Canada) to goal (Portugal) on a rectangular grid. The dynamic programming algorithm of \cref{eqn:DP} is applied to the grid points $i$,$j$ in horizontal and $k$ in vertical direction (not shown). The thick red line is the optimal path.}
    \label{fig:Design_Grid}
\end{figure}

The optimization is based on the well-known dynamic programming~\cite{Bellman_1958DProuting, Bellman_BOOK_DynamicProgramming} technique and extends the implementation by Amb{\"u}hl~\cite{Ambuehl_BolDOr} with the altitude as an additional optimization variable. The working principle can be shown on a basic example, where the goal is to find the shortest distance between Bell Island, Canada and Lisbon, Portugal (\cref{fig:Design_Grid}). In a first step, a three-dimensional grid, connecting the departure and arrival points, is generated. The grid is horizontally divided into $i$ slices of $j$ vertices, and vertically into $k$ levels. Starting from the departure node, the cost (in this example the travel distance) to each subsequent node is calculated and stored. Then, starting from the nodes in the third slice of the grid, the DP algorithm
\begin{equation} \label{eqn:DP}
d_{i,j,k}=\min_{n \in \text{slice}_{i-1}} \left[d_{i-1,n}+\Delta_{i-1,n}^{i,j,k}\right]
\end{equation}
is applied to find the shortest total distance $d_{i,j,k}$ from the departure point to each node of the grid. This is done by minimizing the sum of the a priori known distances $d_{i-1,n}$ and the additional travel distance $\Delta_{i-1,n}^{i,j,k}$. A decision tree consisting of globally optimal sub routes is thus built up which finally reaches the arrival point. The optimal path can then be extracted by going back up the tree from the arrival point. In contrast to this simplified example with a Euclidean distance cost, the real aircraft cost function depends on high-resolution time-varying forecast data and a comprehensive system model. Each path segment thus needs to be simulated using numerical integration, which is a main expansion compared to Amb{\"u}hl~\cite{Ambuehl_BolDOr}.

\subsubsection{Cost Function}

For solar-powered aircraft in particular, a globally optimal point-to-point path is not necessarily the shortest path, but the path with the highest probability of mission success and thus the lowest risk exposure. The cost function mathematically considers risk through \emph{cost terms} which can be grouped as follows:

\begin{itemize}
\item \emph{Flight time}: A low flight time decreases the risk of spontaneous component failure. This mainly becomes of importance if all other costs are small. 
\item \emph{Environmental costs}: The environmental (or meteorological) costs indicate an environmental threat to the airplane. This includes strong wind, wind gusts, humidity, precipitation and thunderstorms.
\item \emph{System costs}: Includes \ac{SoC}, power consumption and the radiation factor, which is the ratio between current solar radiation and clear-sky solar radiation and thus indicates clouds. Flight states with low \ac{SoC}, high power consumption or low radiation factor are avoided by flying at the power-optimal airspeed and evading clouded areas.
\item \emph{Distance to terrain}: Based on a \ac{DEM}, this term helps to avoid terrain collisions while flying in cluttered terrain. The cost term has been extended with respect to our previous work \cite{Wirth_AeroConf2015}.   
\end{itemize}   

\begin{figure}[ht]
    \centering
    \includegraphics[width=0.5\linewidth]{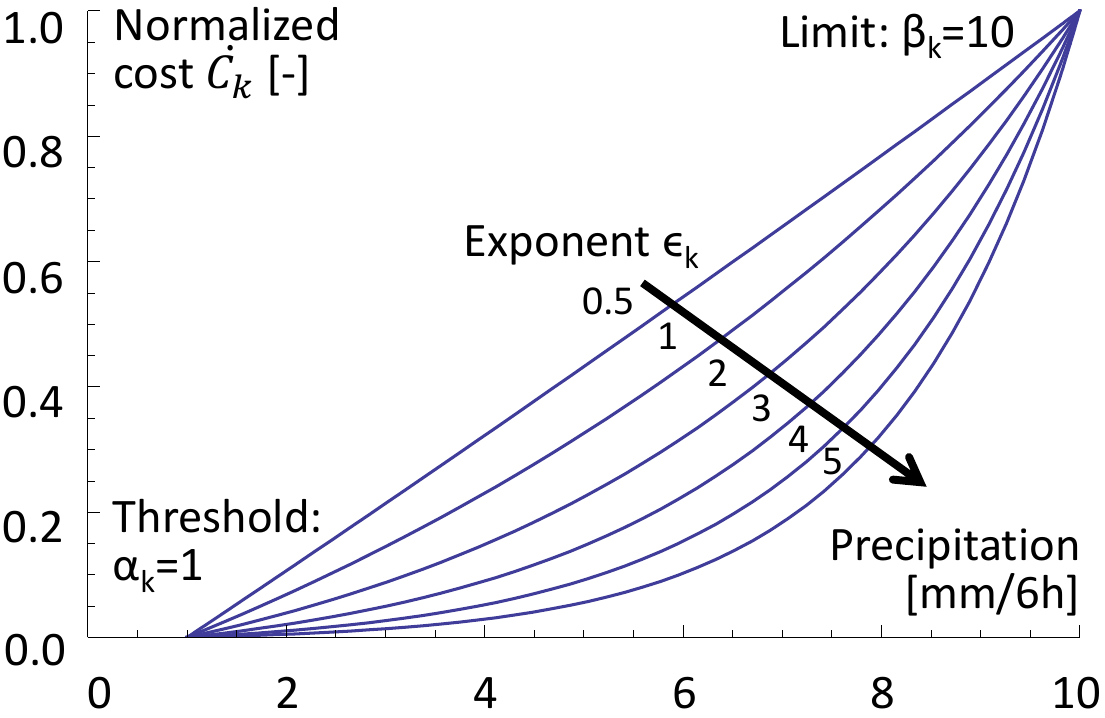}
    \caption{
    Visualization of a cost function and the parameters available to adjust its sensitivity. The precipitation cost is shown as an example.}
    \label{fig:P2P_CostFunction}
\end{figure}

The instantaneous flight time cost is simply the so called flight time cost factor, i.e. $\dot{C}_\text{time}=c_\text{time}$. All other costs $\dot{C}_k$ need to be normalized to allow for a consistent summation and weighting via
\begin{equation}
\dot{C}_k=H(x)\cdot \frac{\exp{(\frac{x_k-\alpha_k}{\beta_k-\alpha_k}\epsilon_k)}-1}{\exp{(\epsilon_k)}-1} \; .
\end{equation}
As illustrated in \cref{fig:P2P_CostFunction}, this normalizes every cost and allows to adjust its influence on the total cost. The parameters $\alpha_k$ and $\beta_k$ define the lower threshold and the upper limit, where the generated cost is bounded. Due to the Heaviside function $H(x)$ values $x_k$ below the threshold generate no cost as they are not in a critical range. Values above the limit are considered too dangerous for the aircraft and thus cause a cancellation of the corresponding path. The exponent $\epsilon_k$ determines the curvature of the cost function.

The accumulated cost for a path segment is finally calculated by summing up all 10 costs to a total cost and integrating it over the flight time, as defined by
\begin{equation} \label{eqn:Design_CostFunctionSum}
C=\int_{t_1}^{t_2}\sum_{k=1}^{10} \dot{C}_k dt \; .
\end{equation}

\subsubsection{Meteorological Forecast Data}

Accurate meteorological data is essential to plan safe and efficient paths for weather-sensitive solar-powered aircraft. The meteorological data used in MetPASS consists of the parameters in \cref{tab:Design_MeteoParams}. Both historical data for mission feasibility analysis and forecast data to pre-plan or re-plan actual missions on site is supported. The data is either obtained from the European Centre for Medium-Range Weather Forecasts (ECMWF) global deterministic IFS-HRES model (horizontal resolution of \unit[0.125]{\degree}) or the European COSMO model (resolution up to \unit[2]{km}). Data time steps range between 1--6 hours and altitude levels between \unit[0]{m} and \unit[1600]{m} above sea-level. The characteristics of all weather data sets are given in \cref{tab:Results_OverviewAndParameters}. The data can be linearly interpolated in time as well as in all three spatial dimensions. More details on weather data import and integration are given by Wirth~\cite{Wirth_Metpass_MasterThesis}.

\begin{table}[htb] 
\caption{Forecast parameters received from ECMWF and COSMO models.} \label{tab:Design_MeteoParams}
\centering
\begin{tabular}{L{3cm} l L{4.8cm} l}
\toprule
 Parameter & Unit & Description & Type \\ 
\midrule
Temperature & \unit{\degree C} & & 3D\\
Relative humidity & \unit{\%} & & 3D\\
Horizontal wind & \unitfrac{m}{s} & Lateral and longitudinal winds& 3D\\
Wind gusts & \unitfrac{m}{s} & Max. wind gust over last time step & 2D\\
Total precipitation & \unit{mm} & Accumulated over last time step & 2D\\
Convective available potential energy & \unitfrac{J}{kg} & Causes updrafts and thunderstorms & 2D\\
Total solar radiation (direct + diffuse) & $\unitfrac{J}{m^2}$ & Accumulated over last time step & 2D\\
Direct solar radiation & $\unitfrac{J}{m^2}$ & Accumulated over last time step & 2D\\
\bottomrule
\end{tabular}

\end{table}


\subsubsection{System Model}
\label{sec:Design_SystemModel}

\begin{figure}[htb]
    \centering
    \includegraphics[width=1.0\linewidth]{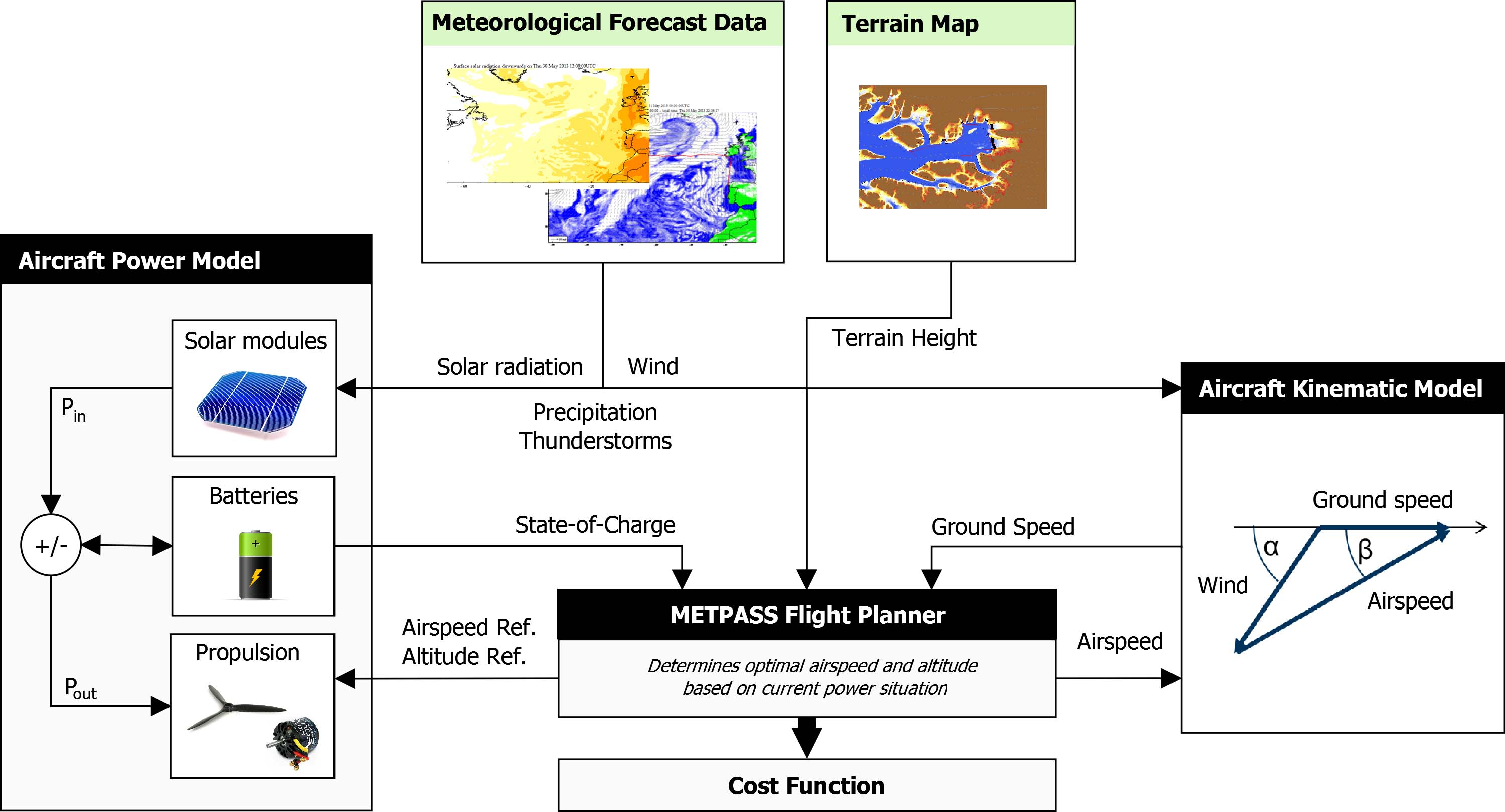}
    \caption{
    The MetPASS system model and the interactions with the Flight Planner that determines the current optimal airspeed and altitude.}
    \label{fig:P2P_SystemModel}
\end{figure}

The system model (\cref{fig:P2P_SystemModel}) simulates the flight along a chosen route and thereby calculates the states required for the cost function. Both the model and the employed weather data are fully deterministic. Given the large mission time scales of hours or even days, aircraft flight dynamics are neglected. The decisive components that are modeled dynamically are the power balance (including time and temperature dependent solar power generation, system power consumption and \ac{SoC}) and the flight kinematics with respect to wind speed and airspeed. 

The flight kinematics are handled by the \code{Aircraft Kinematic Model}, which updates the aircraft position after calculating the wind angle and ground speed. It requires the airspeed, which is determined by the \code{Flight Planner} module --- a representation of the UAV's decision logic --- as a function of the system state. The airspeed may be increased, first, in presence of strong headwind to maintain a certain ground speed or, second, if there is excess solar power available, the battery is already fully charged and the aircraft is not allowed to gain additional altitude. The flight planner can also increase the altitude once $\text{SoC}=\unit[100]{\%}$ either in order to store solar energy into potential energy or to use the wind situation at higher altitude.

The power balance is estimated by the \code{Aircraft System Model}. For the power generation, direct and diffuse solar radiation are considered separately. The incidence angle $\varphi_k$ of the direct radiation is calculated for every solar module $k$ using the solar radiation models presented in \cite{Cornwall_SolarEquations,Fischer_DACHRad,Wirth_Metpass_MasterThesis} under consideration of the aircraft geometry. The radiation onto the surface is then calculated using the cosine of the incidence angle. For the diffuse part of the radiation, the surface is assumed to be horizontal and thus the incidence angle is neglected. With the solar module areas $A_k$ and the efficiencies of the solar modules $\eta_{\text{sm}}$ (which are temperature dependent) and the Maximum Power Point Trackers $\eta_\text{MPPT}$, the total incoming power is
\begin{align}
P_{\text{solar},k} &= \big(I_\text{diff}+I_\text{direct}\cdot\cos(\varphi_k)\big)\cdot A_k\cdot \eta_{\text{sm}} \cdot\eta_\text{MPPT} \\
P_\text{solar} &= \textstyle\sum_k P_{\text{solar},k}
\end{align}
The overall level-flight power consumption of the UAV depends on airspeed $v_\text{air}$ and altitude and thus air density $\rho$. It is generally given by
\begin{equation}
P_\text{level}(\rho,v_\text{air})=\frac{P_\text{prop}(\rho,v_\text{air})}{\eta_\text{prop}(\rho,v_\text{air})}+P_\text{av}+P_\text{pld} \; ,
\end{equation}
where $\nicefrac{P_\text{prop}}{\eta_\text{prop}}$ determines the required electrical propulsion power, and $P_\text{av}$ and $P_\text{pld}$ are avionics and payload power respectively. In our case, the dependence of $P_\text{level}$ on the airspeed $v_\text{air}$ is modeled through
\begin{equation}
P_\text{level}(\rho_0,v_\text{air})=C_2 \cdot v_\text{air}^2+C_1 \cdot v_\text{air} + C_0
\end{equation}
which is identified directly from \emph{AtlantikSolar} power measurement test flights performed at constant altitude and thus air density $\rho_0$. The scaling to different altitudes or air densities is done according to \cite{Pamadi_DynamicsControl_Book} using
\begin{equation} \label{eqn:P_level}
P_\text{level}(\rho,v_\text{air})=\sqrt{\frac{\rho_0}{\rho}} \cdot \left[C_2 v_\text{air}^2 \frac{\rho}{\rho_0} + C_1 v_\text{air} \sqrt{\frac{\rho}{\rho_0}} + C_0\right] \; .
\end{equation}
With regard to the climb rate $\dot{h}$, the total flight power $P_\text{flight}$ is given by
\begin{equation}
P_\text{flight}(\rho,v_\text{air}) = P_\text{level}(\rho,v_\text{air}) + \frac{m_\text{tot} g \dot{h}}{\eta_\text{climb}} \; ,
\end{equation}
with the airplane mass $m_\text{tot}$ and the climbing efficiency $\eta_\text{climb}$. The \ac{SoC} is updated based on the power balance given by
\begin{equation}
\dot{\text{SoC}} = \frac{P_\text{solar}-P_\text{flight}}{E_\text{bat}} \eta_\text{charge} \; ,
\end{equation}
with the total energy of the battery $E_\text{bat}$ and the battery efficiency $\eta_\text{charge}$. Note that $0<\text{SoC}<1$ and a charge rate limit for large SoC's is enforced. More details are presented by Wirth~\cite{Wirth_Metpass_MasterThesis}.

\subsection{Environment-aware Multi-Goal Path Planning}
\label{sec:mg_planning}

\begin{figure}[htb]
    \centering
    \includegraphics[width=0.65\linewidth]{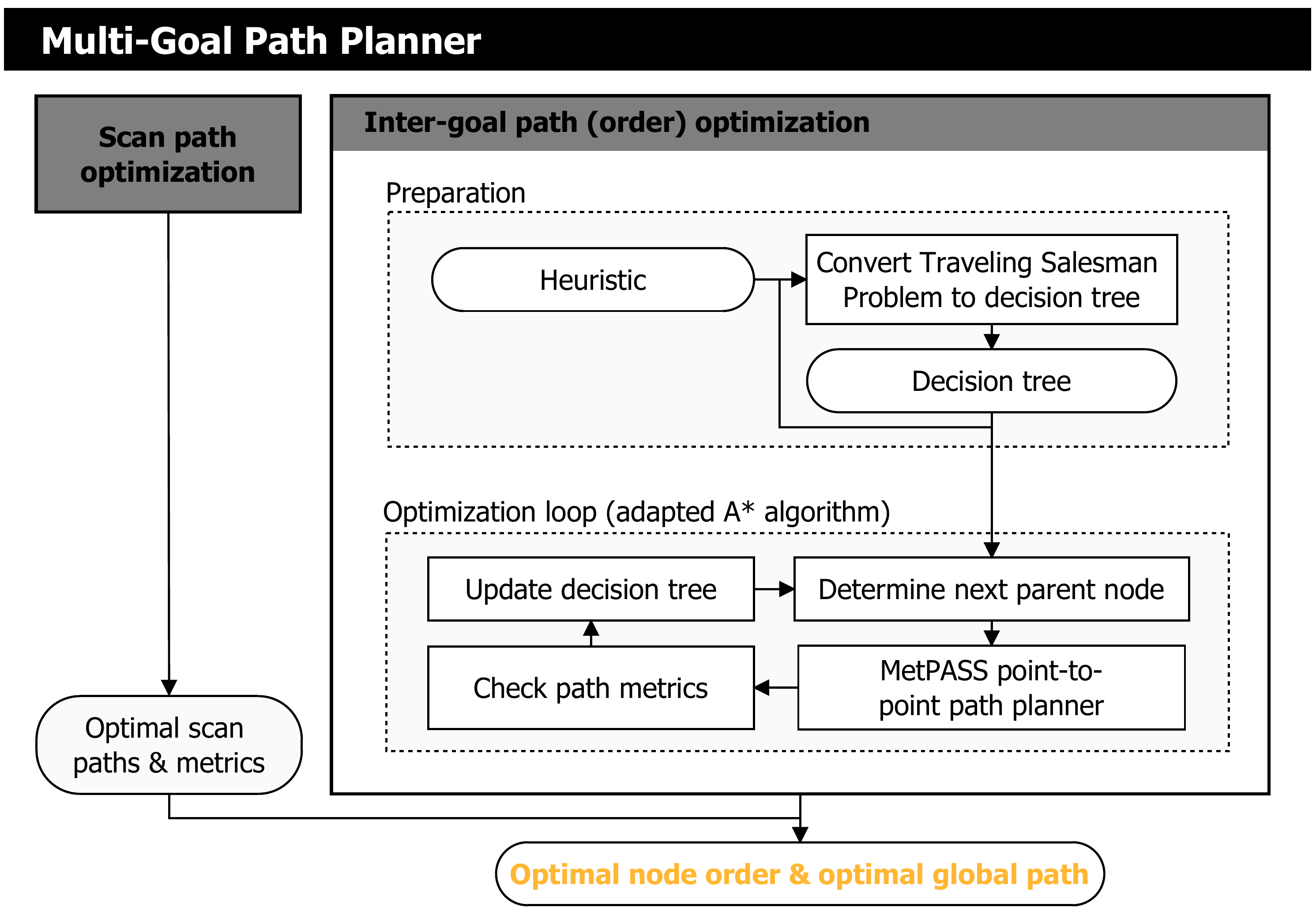}
    \caption{
Enlarged view of the MetPASS multi-goal path planning submodule. The complete MetPASS architecture is shown in \cref{fig:Design_OverviewFlowchart}.
}
    \label{fig:Design_MGOverviewFlowchart}
\end{figure}

The multi-goal path planning problem can be stated as follows: Given fixed start and goal coordinates $S$ and $G$ as well as a set of convex polygonal areas of interest (also called \emph{nodes} $\pazocal{N}$ hereafter), find the order to visit the nodes and thereof derive the optimal path that minimizes the cost function in \cref{eqn:Design_CostFunctionSum} globally over the whole mission. The two sub-functionalities mentioned in \cref{fig:Design_OverviewFlowchart}, i.e. the \emph{scan path optimization} which calculates the scan path inside each area of interest and the \emph{inter goal path planner} which determines the order of and trajectories in between the areas of interest, are visualized in more detail in \cref{fig:Design_MGOverviewFlowchart} and described in the following. 

\subsubsection{Scan Path Optimization}
\label{sec:Design_ScanPathOptimization}

\begin{figure}[htb]
    \centering
    \includegraphics[width=0.45\linewidth]{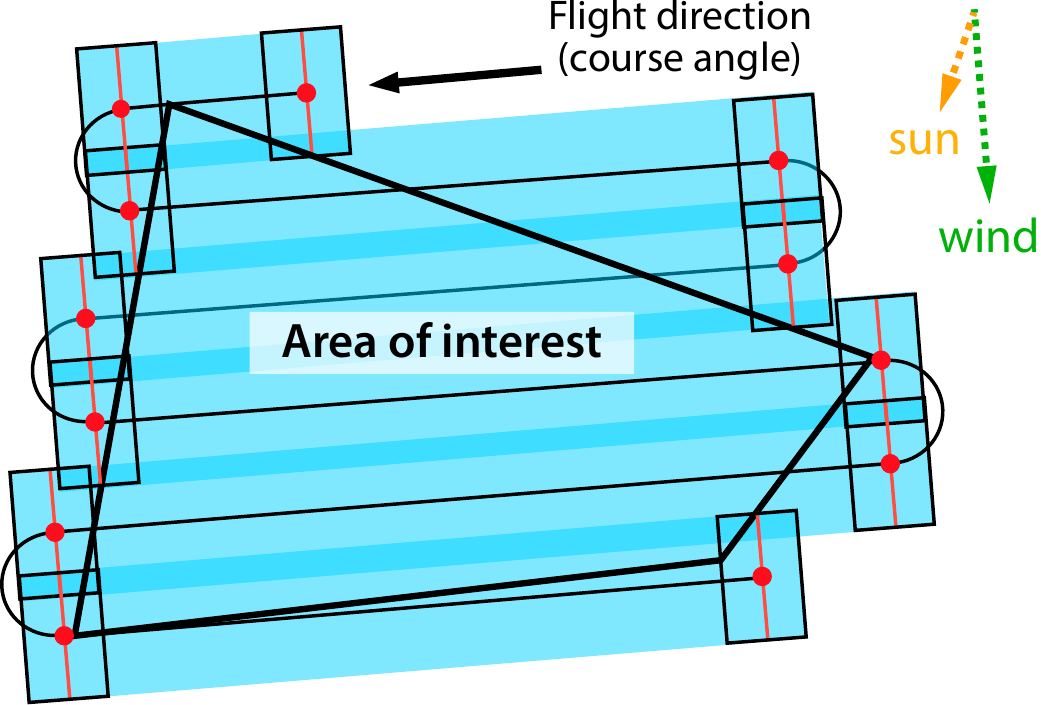}
    \caption{
    Schematic lawn-mower scan path covering an area of interest. The black arrow is the flight direction (course), the black rectangles are selected camera images and the red dots are turn points. Environmental factors such as wind and sun are considered during the optimization.}
    \label{fig:Design_ScanPath}
\end{figure}

The scan path optimization is carried out independently for each area of interest. The problem can be stated as: Given a) the extent of the convex polygonal area of interest, b) fixed scan parameters such as flight altitude, airspeed, camera field of view and desired image overlap, and c) weather data, find a scan path that guarantees complete coverage of the area of interest at minimum cost. For simplicity lawn-mower scan patterns (\cref{fig:Design_ScanPath}), which are widely used in robotics, are employed. 

The main parameter that needs to be optimized is the course angle. For constant wind, both simple geometric calculations and the simulations in \cite{Foerster_Metpassv2} show that for lawn-mower patterns with equal outward/backward-distances a course perpendicular to the wind direction results in minimum flight time. However, first, in our case the wind is not constant in space, and second, the polygon can be of arbitrary but convex shape such that the outward/backward-distances can differ significantly. Third, we are interested in a minimum total cost instead of a minimum flight time path: For example, the sun position and the resulting solar power income over all course angles need to be considered. 

The course angle therefore can not be calculated analytically, but it has to be optimized via simulating scan paths over a range of course angles. Both the straight lines and the 180{\degree}-turns of the lawn-mower pattern are simulated using the system model of \cref{sec:Design_SystemModel}. The optimal polygon corner to start from is found in the same process. The combination of course angle and polygon start corner with minimum cost is selected as the solution. The details of the optimization process are presented in \cite{Foerster_Metpassv2}. The final results are the optimized scan path (\cref{fig:Design_ScanPath}) and the scan metrics path length, flight time, cost, and change in state of charge $\Delta \text{SoC}$. These metrics are required inputs for the inter-goal path planner.


\subsubsection{Inter-Goal Path Optimization}
\label{sec:Design_InterGoalPathOptimization}

The inter-goal path optimization finds the optimal order to visit all areas of interest and thereof derives the overall path between them and the base. When aiming only for minimum flight distance, then the trivial circular path shown in \cref{fig:Intro_Collage} is likely a good solution to this well known Traveling Salesman Problem (TSP, \cite{Hoffman_TSP}). Instead of minimum distance our goal is minimum path cost. In addition, weather changes make this problem time-dependent. The problem is also asymmetric because, for example because of wind, flying from node A to node B can cause a different cost than flying from B to A. We are therefore trying to solve a Time Dependent Asymmetric Traveling Salesman Problem (TDATSP). Both the TSP and TDATSP are NP-hard problems and thus computationally expensive to solve. For example, the brute-force approach to solving the standard TSP is of order $O(N!)$, where $N$ is the number of nodes excluding the start/goal node. For our applications we mostly have $N<10$, however, each calculation of a point-to-point path requires MetPASS to solve a dynamic programming problem that takes between 3 seconds and 10 minutes for typical grid resolutions and distances (\cref{tab:Results_ComputationTimes}). The algorithms used to solve the TDATSP should therefore aim to decrease the required edge cost calculations.

\paragraph{Solving the time-dependent asymmetric traveling salesman problem}

A number of algorithms to solve the time dependent asymmetric traveling salesman problem exist. Dynamic programming approaches were introduced early-on by Bellman~\cite{Bellman_DPforTSP} for the TSP. More recently, Malandraki introduced an optimal DP approach \cite{Malandraki_DPforTDATSP} and a faster restricted \cite{Malandraki_restrDPforTDATSP} DP method for solving the TDATSP. However, these were not selected because the former requires the computation of all edge costs, while the latter does not guarantee optimality. Similarly, the Simulated Annealing approaches introduced by Schneider~\cite{Schneider_TDATSP} do not provide optimality guarantees. The genetic algorithms introduced for solving the TDATSP by Testa et al.~\cite{Testa_TDATSP} tend to provide lower computational performance than the DP methods. They are asymptotically optimal, i.e. cannot guarantee to find the optimal solution in finite time. 

\begin{figure}[htb]
    \centering
    \includegraphics[width=1.0\linewidth]{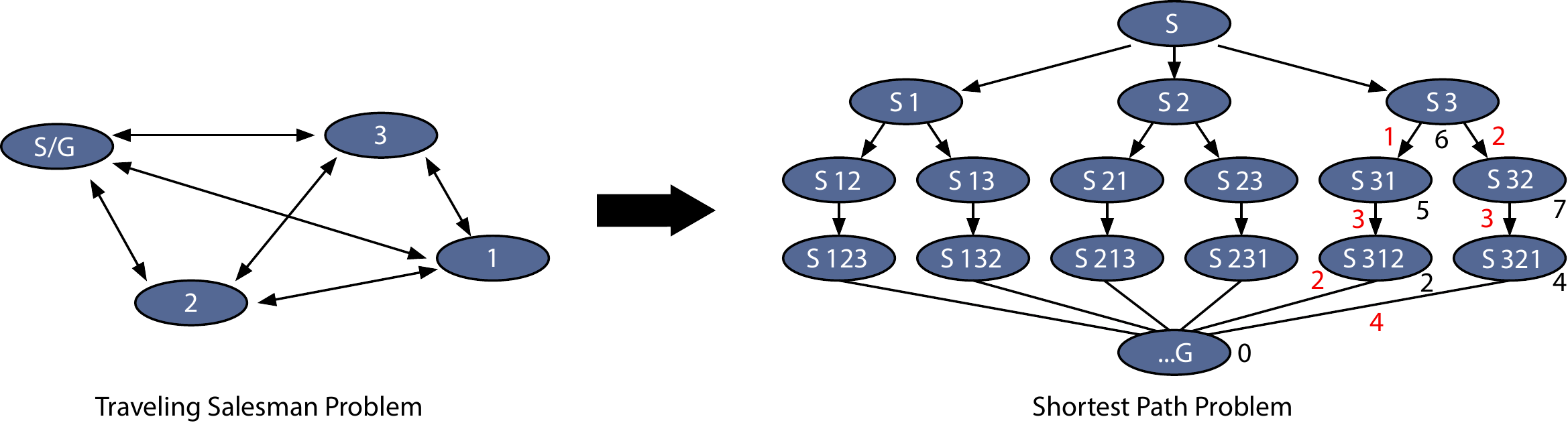}
    \caption{
    Traveling salesman problem and the equivalent shortest path problem that label correcting methods can be applied to. With the notation of this paper, i.e. excluding the start/goal in the node count and only the start vertex in the vertex count, the TSP has $N=3$ nodes and the SPP graph has $V=21$ vertices. The SPP graph contains exemplary point-to-point heuristics (red) and cost-to-go heuristics (black).}
    \label{fig:Design_ConversionTSPtoSPP}
\end{figure}

Finally, efficient label correcting methods can be applied if the TDATSP is converted into a Time Dependent Shortest Path Problem (TDSPP) as described by Bertsekas~\cite{Bertsekas_Book_DPandOptCtrl}. This approach was chosen for this paper. The resulting tree is visualized in \cref{fig:Design_ConversionTSPtoSPP}. While the original TSP graph contains the $N=3$ nodes $\pazocal{N}=\{1,2,3\}$ excluding the start/goal nodes, the SPP graph contains the $V=21$ vertices $\pazocal{V}=\{S1, S2, S3, S12, \dots, S312G, S321G\}$ excluding the start vertex $S$ but including those vertices\footnote{Note that in the following discussions of the graphs, the indices $n,m\in\pazocal{N}$ always refer to nodes while the indices $v,w\in\pazocal{V}$ always refer to vertices.} that finish with the goal node $G$. The root of the SPP graph represents our start node $S$, and every branch is a valid (though not necessarily feasible or even optimal) solution path that covers all areas of interest and ends at the start node (called goal $G$ for clarity) again. The cost for traversing an edge in the new graph, i.e. for going from vertex $v$ to a child vertex $w$, is equal to the cost of going from the node $n$ represented by the last digit in the parent label to the node $m$ represented by the last digit in the child label. These costs are time dependent. Finally, of the many well-known methods that can be applied to the TDSPP graph, a modified version of the A*-algorithm introduced by Hart et al.~\cite{Hart1968AStar} was selected. When used with a properly designed heuristic, A* can guide the search and can thereby significantly reduce the amount of required edge cost calculations while still providing globally optimal paths.

\paragraph{Heuristic calculation}

For every vertex $v \in \pazocal{V}$ in the SPP graph of \cref{fig:Design_ConversionTSPtoSPP}, the heuristic $h_v$ describes an estimate of the cost to fly to the goal $G$ via all yet unvisited nodes. The heuristic has to be a lower bound on the path cost to avoid discarding optimal paths and to thus retain the optimality guarantees of the A*-algorithm~\cite{Hart1968AStar}. However, the closer it is to the actual optimal path cost the larger the reduction in edge cost evaluations will be. The calculation of the heuristic is performed in two steps:
\begin{itemize}
\item First, a lower-bound cost estimate $h_{nm}$ for a path from TSP node $n$ to node $m$ is computed\footnote{Note that $h_{nm}$ can be calculated between the nodes $\pazocal{N}$ of the TSP graph instead of the vertices $\pazocal{V}$ of the SPP graph because the point-to-point heuristic is time-independent.}. As for the standard edge cost calculation this estimate is calculated using \cref{eqn:Design_CostFunctionSum}, however, for the heuristic we assume that a) the path is the straight line (or orthodrome) between nodes $n$ and $m$, b) we start with full batteries and c) have optimal weather conditions (e.g. tail wind, low cloud cover and precipitation, etc.) during the whole flight. The optimal weather conditions are searched over all grid points of the standard DP point-to-point mesh (\cref{fig:Design_Grid}) and all times within a user-specified time horizon.
\item Second, the cost-to-go heuristic $h_v$ is calculated for each vertex $v\in\pazocal{V}$ by summing up the point-to-point cost estimates $h_{nm}$ from the SPP graph's lowest level vertices (which represent complete round-trips and are assigned a heuristic value of zero) inside each tree to vertex $v$. 
\end{itemize}

\paragraph{Optimization loop with an extended A* algorithm}

The A*-algorithm shown in \cref{alg:Design_AStar} uses the following notation: The current lowest cost to go from $S$ to vertex $v\in\pazocal{V}$ is $g_v$ and the cost to go from vertex $v$ to vertex $w$ is $c_{vw}$, where $w$ is a successor of $v$ and $p_v$ is the parent vertex of $v$ on the shortest path from $S$ to $v$ found so far. In contrast to the standard algorithm by Hart~\cite{Hart1968AStar}, the implementation in MetPASS needs to be extended to consider 
\begin{itemize}
\item \emph{The time dependence} of the edge costs due to changing weather. Each vertex $v$ is therefore assigned a departure time $t_v$.
\item \emph{The \ac{SoC} dependence} of the edge costs. Each vertex $v$ is assigned a departure state of charge $\text{SoC}_v$. The edge costs $c_{vw}$ between the vertices already include both the \ac{SoC} and time dependency, i.e. $c_{vw}=c_{ij}(t_v,\text{SoC}_v)$.
\item \emph{The scan path metrics}, i.e. the time and state of charge changes $T$ and $\Delta\text{SoC}$ when inspecting an area of interest. Note that both are assumed time-independent for the optimization. The scan costs are thus the same for every branch in the SPP Graph and don't need to be considered in the optimization. This speeds up the optimization significantly. However, it also means that if the ratio between the time spent on the scan paths and the time spent on the inter-goal paths exceeds a certain value, then the global multi-goal path looses its optimality properties.
\end{itemize}
The extended A* algorithm starts with the initialization, in which all vertices are labeled as ``unlabeled'' and all $g_k$ where $k\in\pazocal{V}$ are set to $\infty$. Only the start vertex is labeled ``open" and $g_S$ is set to $0$. Time and \ac{SoC} are initialized to user-specified defaults. Next, in each iteration, the algorithm picks a vertex to process by finding amongst all vertices that are labeled ``open" the one with the lowest total expected cost $g_v+h_v$. When a vertex $v$ (from now on called ``parent" vertex) is selected, the following two equations are checked for each child $w$ of that vertex (note again that the time and \ac{SoC} dependency of the edge costs are already incorporated in $c_{vw}$):
\begin{align} \label{eqn:AStar_Checks}
g_v+c_{vw} &< g_w \\
g_v+c_{vw}+h_w &< g_G
\end{align}
If the two checks are passed, $g_w$ and $p_w$ in the decision tree are updated and $w$ is labeled as ``open". In addition to this standard A*-behaviour, our algorithm also updates the time $t_w$ and state of charge $\text{SoC}_w$ at the vertex using the scan path metrics $T_w$ and $\Delta\text{SoC}_w$ (which are both time-independent and can thus be replaced by the $T_m$ and $\Delta\text{SoC}_m$ of the corresponding node). Otherwise, nothing happens and the next successor vertex is considered. As soon as all successor vertices of the current parent vertex have been processed, the parent vertex is labeled as ``closed" and the next iteration is initiated. The algorithm is terminated as soon as the terminal node $G$ is part of the label of the selected parent vertex.

\begin{pseudocode}[ruled]{Extended A*-algorithm used in MetPASS}{ }
\label{alg:Design_AStar}
\text{\textbf{comment:} Finds shortest path between start $S$ and goal $G$. Note}\\ 
\hspace{13.4ex}\text{that the edge cost $c_{vw}$ and the travel time $t_{vw}$ always}\\
\hspace{13.4ex}\text{incorporate the dependency on $(t_v,$SoC$_v)$}\\
g_S \GETS 0\\
l_S \GETS \text{"open"}\\
t_S \GETS 0\\
\text{SoC}_S \GETS \text{initial SoC specified by user}\\
g_k \GETS \infty \quad \forall k \in V\\
l_k \GETS \text{"unlabeled"} \quad \forall k \in V\\
v \GETS S\\
\WHILE G\notin v \DO
\BEGIN
	\FOREACH w \in \text{children}(v) \DO
	\BEGIN
		\IF g_v + c_{vw} < g_w \AND g_v + c_{vw} + h_w < g_G
		\THEN 
		\BEGIN
			g_w \GETS g_v + c_{vw}\\
			l_w \GETS \text{"open"}\\
			t_w \GETS t_v + t_{vw} + T_w\\
			\text{SoC}_w \GETS \text{SoC}_v + \Delta \text{SoC}_{vw} + \Delta \text{SoC}_w\\
			p_w \GETS v
		\END
	\END\\
	l_v \GETS \text{"closed"}\\
	v \GETS \text{"open" node with smallest value } g_k + h_k
\END
\end{pseudocode}

Note that even after the adapations, this implementation of the A* algorithm is still guaranteed to find the optimal multi-goal path under the assumption that the areas of interest are small compared to the total path length. First, for the A* algorithm to be valid, the principle of optimality in the direction of execution (forward in our case) has to hold~\cite{Hamacher2006}. The forward principle of optimality holds for the optimal path $P_{SG}(t)$ from $S$ to $G$ if and only if for an arbitrary intermediate vertex $v$, the optimal path $P_{Sv}(t)$ from $S$ to $v$ is contained in $P_{SG}(t)$. This does not hold for arbitrary paths~\cite{Hamacher2006}. However, as our TDSPP graph is an \emph{arborescence} (a special case of a \emph{polytree}), there is only a single path from the root $S$ to any other vertex, and the forward principle of optimality thus holds. Second, many comparable extensions of the A* algorithm for time-dependent shortest path problems~\cite{Chabini2002} require satisfying the first-in-first-out (FIFO) property $t_1\leq t_2 \implies t_1+t_{vw}(t_1)\leq t_2+t_{vw}(t_2)$. In our case this is not required because the graph is a polytree and waiting at nodes is not allowed. In addition, while the scan paths are considered time independent during the inter-goal optimization, they are afterwards recalculated based on current weather data to have a physically correct state and cost prediction in the planned path.

\section{Implementation, Verification and Preliminary Results}
\label{sec:Implementation}

\subsection{Implementation}
\label{sec:Impl_Impl}

\subsubsection{Overview}
\label{sec:Impl_Impl_Overview}

\begin{figure}[htbp]
    \centering
    \includegraphics[width=1.0\linewidth]{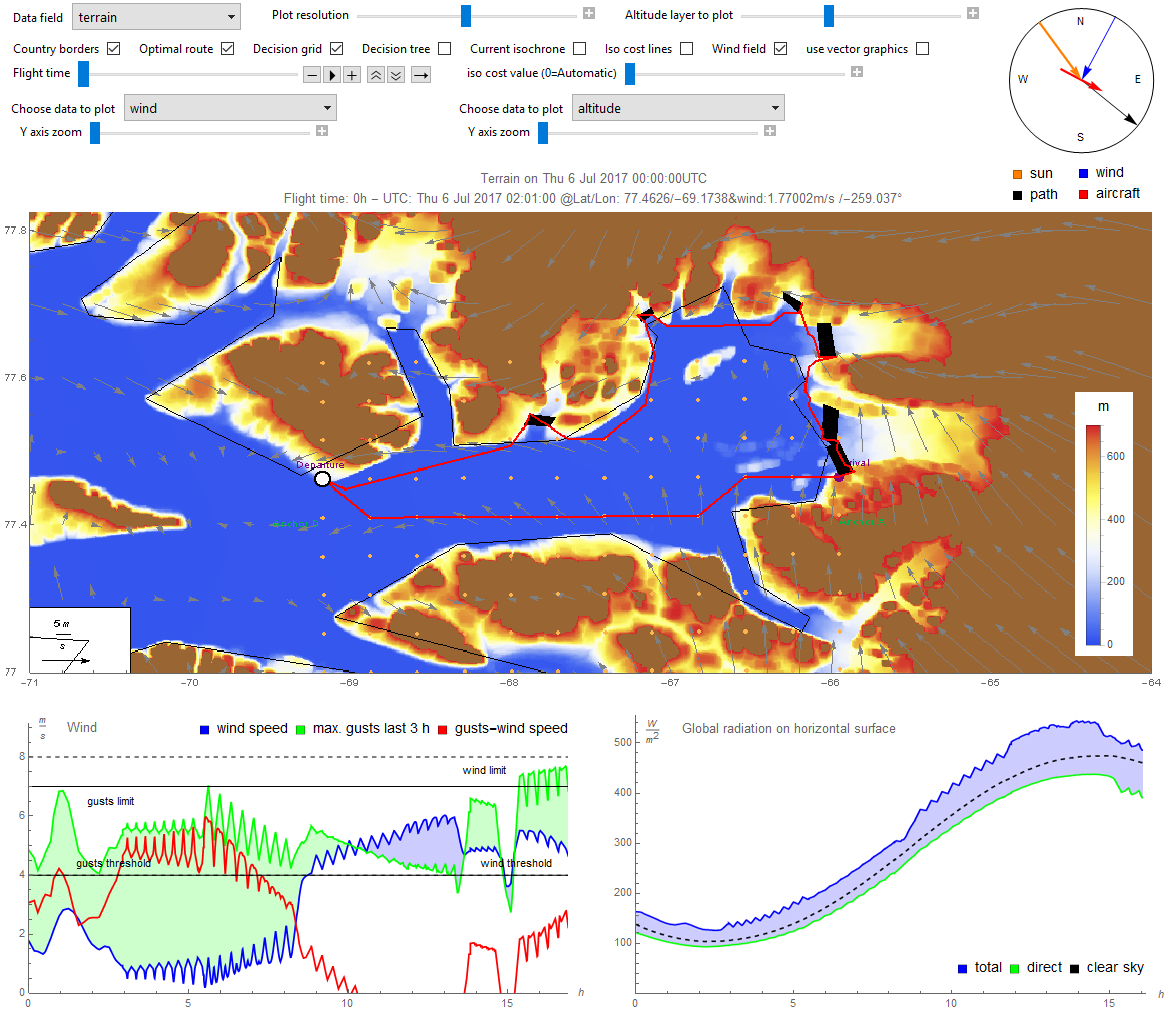}
    \caption{
    MetPASS Graphical User Interface. Top left: Visualization controls used to configure all plots. Top right: Sun, wind and path angle indicator. Center: Main visualization with the areas of interest (black), proposed path (red), current aircraft position (white dot), a terrain altitude map (color bar) and wind vector overlay (grey). Bottom: Expected wind speeds/gusts and solar radiation at the aircraft position over the full flight.}
    \label{fig:Impl_GUI}
\end{figure}

MetPASS is not only an abstract trajectory planner, but provides a comprehensive yet easy-to-use software environment for large-scale environment-aware mission planning and analysis. The Graphical User Interface (GUI) shown in \cref{fig:Impl_GUI} supports the in-detail analysis of the path planning decisions. This is important because especially high-performance solar-powered UAVs are usually fragile and expensive, and the operator therefore needs to have maximum confidence in the flight path. Using MetPASS usually involves the following steps:

\begin{itemize}
\item \emph{Setup} of the aircraft and inspection mission via standard text files
\item \emph{Pre-inspection} of the overall meteorological situation through the GUI
\item \emph{Flight path planning} via the algorithms of \cref{sec:Design}
\item \emph{Post-processing and visualization} of the flight path, the terrain, the environmental risks (rain, wind, thunderstorm, etc.) and the system state (power income and consumption, state of charge, speeds) via the GUI.
\item \emph{Upload to UAV:} The path's waypoints can be exported and uploaded to the vehicle autopilot via any compatible ground control station\footnote{e.g. QGroundControl, \url{http://www.qgroundcontrol.org}}.
\end{itemize}

Note that due to its fully parametrized system model, MetPASS can not only be used with solar-powered UAVs but with any battery-powered aircraft. For a typical mission, it will usually be leveraged for, first, mission feasibility analysis using historical data, second, to then pre-plan the actual mission using forecasted weather data (including finding the optimal launch time in a certain launch window) and, third, to re-plan the waypoints (including waypoint upload) once updated weather forecasts have become available during the flight.

\subsubsection{Performance Optimization}

MetPASS is implemented in \emph{Wolfram Mathematica}. To reduce the computation time for complex missions MetPASS integrates a number of performance optimizations: First, previous calculation results, e.g. the heuristics or pre-planned scanned paths, are cached and automatically re-used. Second, compiled C-code instead of much slower Mathematica-code is used for the computationally expensive DP-based point-to-point path planner. Third, parallel computation with theoretically up to $j\cdot k$ cores ($j$ vertices, $k$ altitude levels, see \cref{fig:Design_Grid}) is supported. 
In addition, more fundamental approaches to save edge cost evaluations, such as discretizing edge costs both with respect to time as well as state of charge, are integrated. However, while these are interesting concepts, F{\"o}rster \cite{Foerster_Metpassv2} found that they only yield negligible improvements in computational performance. They are thus not further elaborated on here.

\subsection{Validation and Preliminary Results}
\label{sec:Impl_Validation}

This section presents preliminary results to validate the most fundamental level of the multi-goal mission planning capability, i.e. the MetPASS point-to-point planner. To independently verify the system model and the optimization with respect to individual cost components, first, \emph{unit tests} that only use a subset of cost components and environmental parameters are performed. Second, several costs or environmental parameters are combined and altitude decision making is added. Third, all cost components and environmental effects are combined and the individual cost parameters are tuned using historical test cases. Departure and arrival points are the same as in \cref{fig:Design_Grid}. More details on the validation process are presented by Wirth~\cite{Wirth_Metpass_MasterThesis}.
 
\paragraph{Unit tests}

\begin{figure}
\centering
\includegraphics[width=\textwidth]{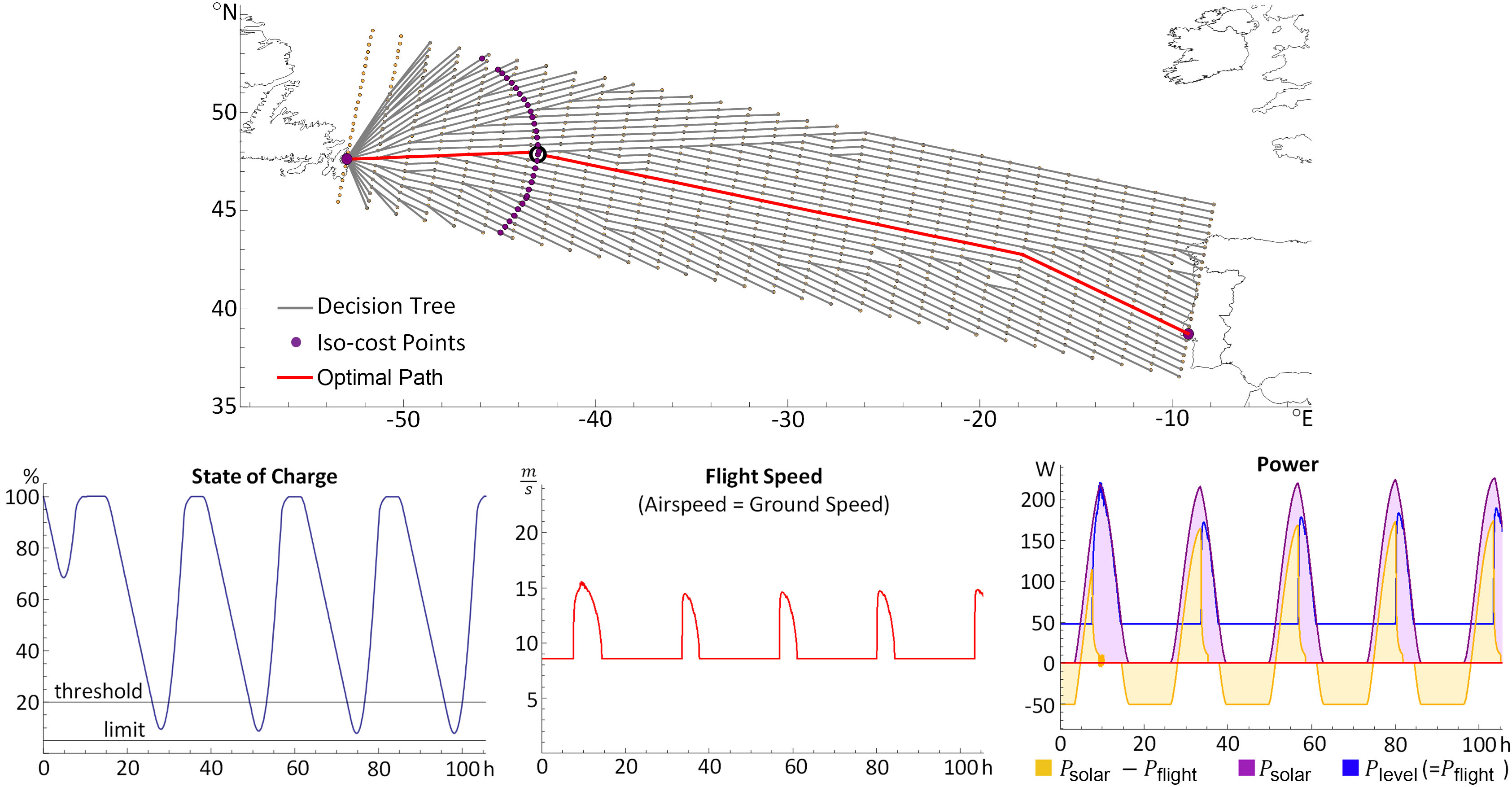}
\caption{First unit test: 2D, i.e. no altitude changes allowed, only time costs, default weather parameters. The iso-cost points form a circle around the start point. \ac{SoC} and power income fluctuate with the day/night cycle. The airspeed is increased when $\text{SoC}=\unit[100]{\%}$ and excess solar power is available.}
\label{fig:Impl_UnitTest1}
\end{figure}

In the first unit test, the required flight time is the only active cost. All forecast parameters are set to default values (no wind and clear-sky solar radiation). \Cref{fig:Impl_UnitTest1} shows that the dynamic programming algorithm finds the fastest path (an orthodrome projected onto the grid) successfully and the system model produces the expected results. The flight planner increases the airspeed as soon as the battery is fully charged and excess solar power is available. The flight time under these zero-wind conditions is about \unit[106]{hours}, the total distance is \unit[3650]{km}.

\begin{figure}
\centering
\includegraphics[width=\textwidth]{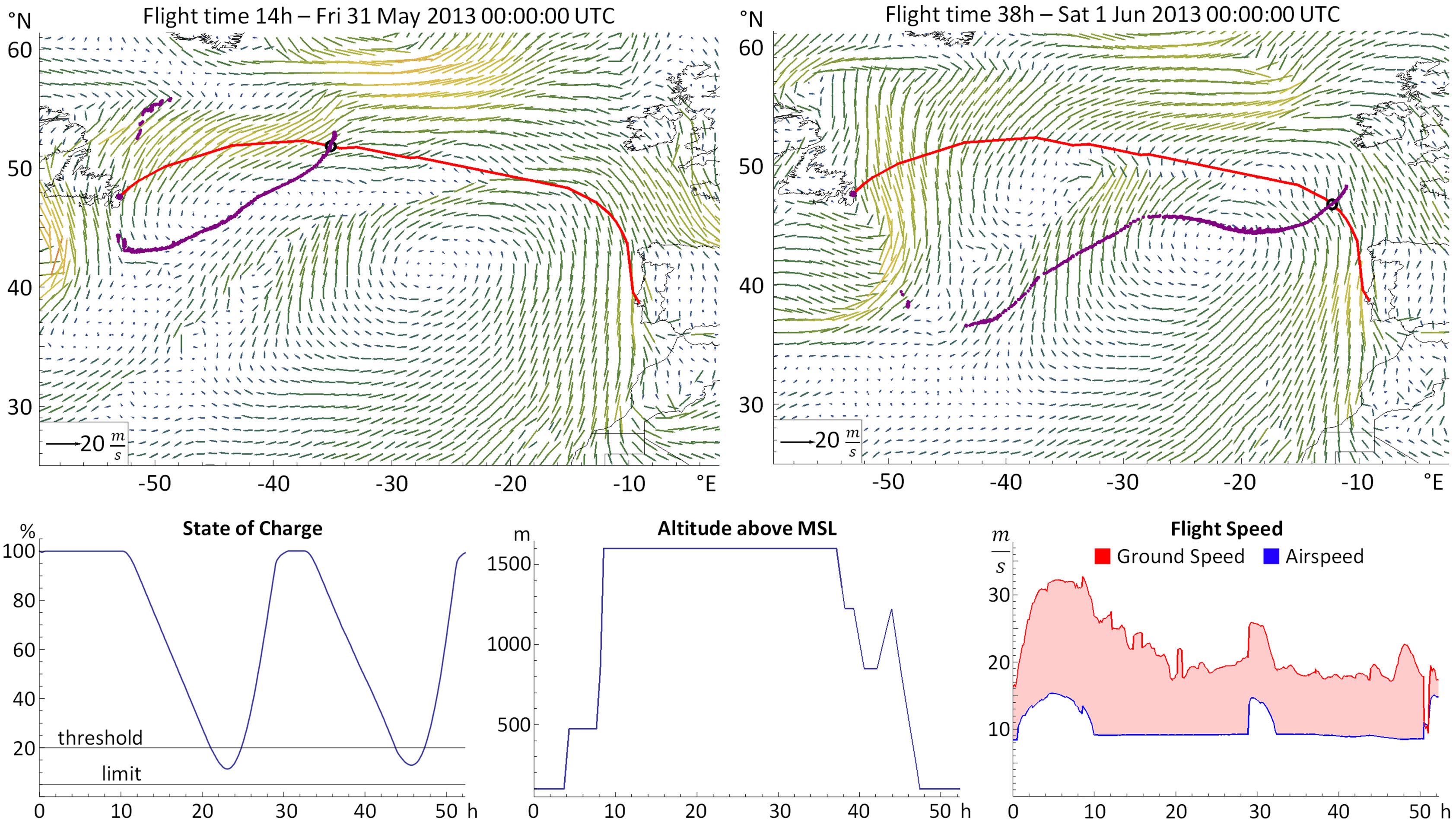}
\caption{Second unit test: 3D, i.e. altitude changes allowed, only time costs, wind conditions taken into account. The flight time is minimized by exploiting winds (including those at higher altitude) such that $v_\text{gnd}>v_\text{air}$. The flight time reduces from \unit[106]{hours} without wind to only \unit[53]{hours}.}
\label{fig:Impl_UnitTest2}
\end{figure}

In the second unit test, the flight time is still the only active cost, but the ECMWF wind forecasts are now considered. \Cref{fig:Impl_UnitTest2} shows the optimal path with departure on May 30, 2013 10:00 UTC. The three-dimensional path optimization under time-varying wind conditions can be observed. The planner optimizes the altitude to leverage the stronger winds at higher altitude such that, overall, $v_\text{gnd}>v_\text{air}$. The flight time reduces from \unit[106]{hours} without wind to only \unit[53]{hours}.

\begin{figure}
\centering
\includegraphics[width=\textwidth]{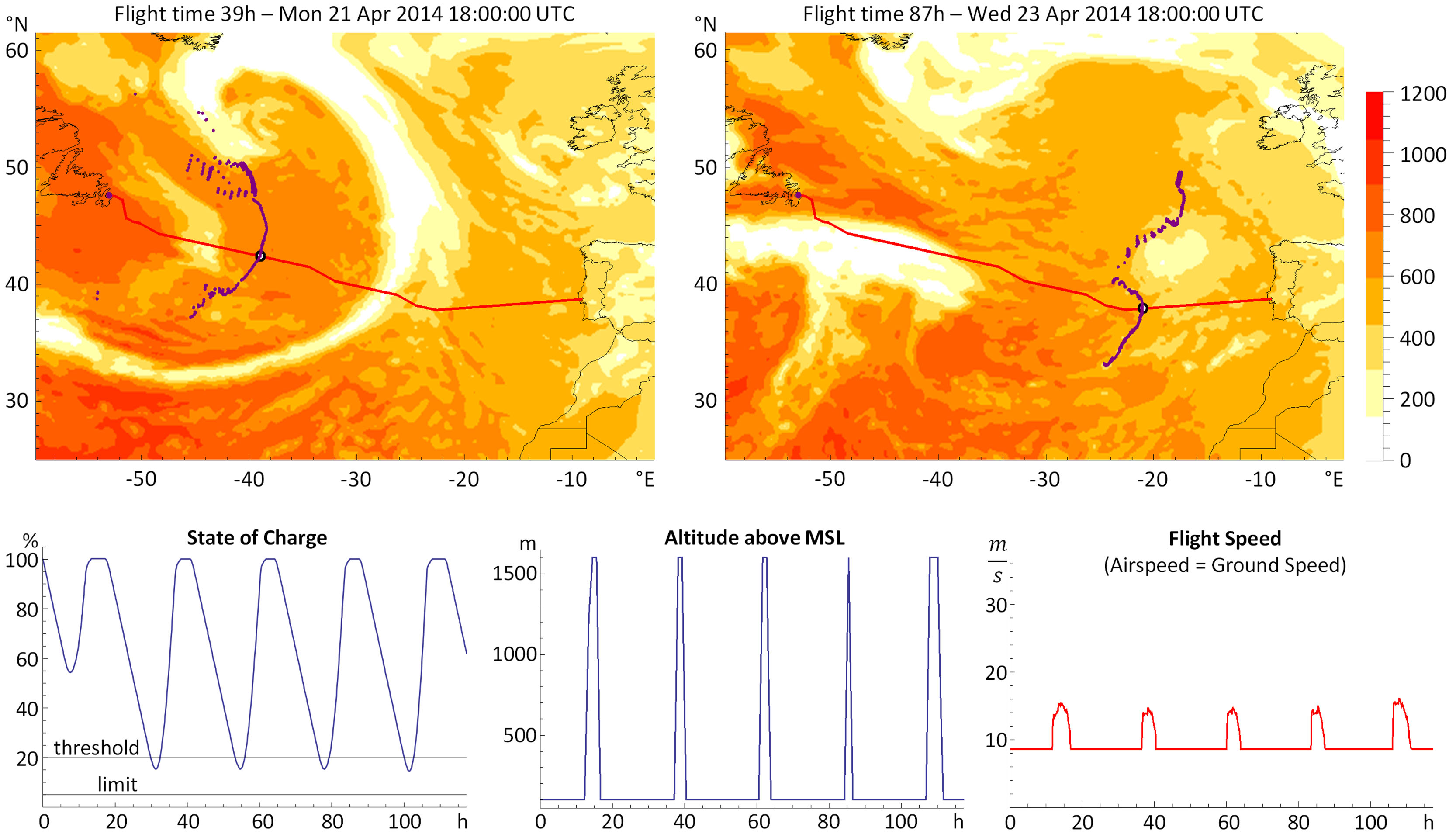}
\caption{Third unit test: The costs are solar radiation, power consumption and flight time. Only the radiation weather parameter is applied. Both altitude and airspeed are increased when $SoC=\unit[100]{\%}$ and excess solar power is available. The iso-cost points illustrate the influence of the radiation.}
\label{fig:Impl_UnitTest3}
\end{figure}

A unit test combining only power balance costs, namely solar radiation, power consumption and flight time, is illustrated in \cref{fig:Impl_UnitTest3}. The applied meteorological parameters are only direct and total solar radiation. The optimal path follows highly radiated areas during the day and the shortest distance at night. The altitude changes show that the power and flight time cost interaction works as desired: At night, the lowest altitude is chosen to minimize the power consumption. During the day, once $SoC=\unit[100]{\%}$ and excess solar power is available, the top altitude level is chosen to store potential energy and to increase the airspeed at a given power consumption.

\paragraph{Parameter refinement}

After successful unit testing, the cost parameters were refined to balance the influence of individual costs on the overall planning outcome. The process starts with estimated initial values that were either recommended by meteorologists or by the aircraft operators. Then, multiple historical test cases were used to iteratively fine tune the parameters. Details on the whole process are available in \cite{Wirth_Metpass_MasterThesis}. As shown in \cref{sec:Results}, the final cost parameter sets (\cref{tab:Results_OverviewAndParameters}) allow the planner to exploit advantageous winds and regions with low cloud cover.

\section{Results}
\label{sec:Results}

Within the context of ETH Zurich's solar-powered UAV development efforts, MetPASS was used to plan a demanding set of missions that could later often be executed successfully by ETH's \emph{AtlantikSolar} UAV. \Cref{tab:Results_OverviewAndParameters} presents an overview over these missions and the main parameters. First, \emph{AtlantikSolar}'s 81-hour continuous solar-powered loitering mission \cite{Oettershagen_JFR2017} that represents the current world record in flight endurance for aircraft below \unit[50]{kg} mass is described. This flight is mainly used to verify the MetPASS system model. Second, results from planning the first-ever autonomous solar-powered flight over the Atlantic are presented. A feasibility analysis is derived. Third, results from planning two large-scale multi-goal inspection missions for glacier monitoring above the Arctic Ocean are discussed. \emph{AtlantikSolar} performed such a flight over Greenland in summer 2017.

\begin{table}[htbp] 
\caption{Overview over the missions planned using MetPASS. The airplane parameters differ because three UAV versions (AS-1, AS-2 and AS-3) are used: $\eta_\text{sm}$ and $e_\text{bat}$ improve over time (but $e_\text{bat}$ was decreased for the Arctic because of the low temperatures), and $P_\text{flight}$ only increases because payload (e.g. cameras, satellite communication) was added. The cost parameters also vary: For example, the Atlantic mission (which was planned first, i.e. in 2014) had higher wind thresholds than the other two missions because constant winds were not considered as dangerous over the open ocean, much lower \ac{SoC} limits because the AS-1 UAV performance was lower, and a lower time cost factor to put more emphasis on being safe rather than fast. In the Arctic missions, (A) represents Bowdoin and (B) represents the six-glacier mission.} \label{tab:Results_OverviewAndParameters}
\centering
\begin{tabular}{l l l l}
\toprule
 \multirow{2}{*}{\textbf{Mission}} & \textbf{81h-flight} & \textbf{Atlantic} & \textbf{Arctic} \\
 & (Sec. \ref{sec:Results_81hFlight})& (Sec. \ref{sec:Results_Atlantic}) & (Sec. \ref{sec:Results_Arctic_Bowdoin}/\ref{sec:Results_Arctic_TheHeilprins})\\
\midrule

\multicolumn{4}{c}{\emph{Path, grid and simulation parameters}}\\
Mission type & Loitering & Point-to-point & Multi-goal\\
Grid points (LxWxH) & 1x1x1 & 40x113x5 & A:30x25x1, B:12x9x1 \\
Altitude range (MSL)& \unit[600]{m} & \unit[100--1600]{m} & \unit[800]{m} \\
Simulation time step& \unit[600]{s} & \unit[600]{s} & \unit[600]{s} \\[0.8ex]

\multicolumn{4}{c}{\emph{Meteorological parameters}}\\
Data type & Forecasts & Historical & Forecasts \\
Model type & COSMO-2 & ECMWF HRES & ECMWF HRES \\
Long. resolution & \unit[2]{km} & 0.125\degree & 0.2\degree \\
Lat. resolution & \unit[2]{km} & 0.125\degree & 0.1\degree \\
Time resolution & \unit[1]{h} & \unit[6]{h} & \unit[3]{h}\\[0.8ex]

\multicolumn{4}{c}{\emph{Airplane parameters}} \\
Aircraft (Year) & AS-2 (2015) & AS-1 (2014) & AS-3 (2017)\\
$m_\text{tot}$ & \unit[6.9]{kg} & \unit[7.0]{kg} & \unit[7.4]{kg} \\
$m_\text{bat}$ & \unit[2.92]{kg} & \unit[2.92]{kg} & \unit[2.92]{kg} \\
$e_\text{bat}$ & \unitfrac[240]{Wh}{kg} & \unitfrac[230]{Wh}{kg} & \unitfrac[222]{Wh}{kg}\\
$\eta_\text{sm}$ & \unit[23.7]{\%} & \unit[20.0]{\%} & \unit[23.7]{\%}\\
$P_\text{flight}(v_\text{air}^\text{opt})$ & \unit[42]{W} & \unit[47]{W} & \unit[57]{W} \\[0.8ex]

\multicolumn{4}{c}{\emph{Cost parameters}} \\
\begin{tabular}[t]{@{\hskip0pt} l @{\hskip0pt}} 
\\
State of charge \\ 
Radiation factor\\ 
Exc. power cons.\\
CAPE \\
Wind \\
Wind gusts \\
Precipitation \\
Humidity \\
Altitude AGL\\
\end{tabular}
& 
\begin{tabular}[t]{@{\hskip0pt} l @{\hskip7pt} l @{\hskip7pt} l @{\hskip2pt}}
$\alpha$ & $\beta$ & $\epsilon$ \\ 
\hline
0.4 & 0.2 & 3 \\
0.8 & 0.05 & 3 \\
0 & 200 & 1 \\
100 & 2000 & 3 \\
6 & 12 & 3 \\
9 & 15 & 3 \\
0.1 & 10 & 3 \\
80 & 100 & 5 \\
- & - & - \\
\end{tabular}
&
\begin{tabular}[t]{@{\hskip0pt} l @{\hskip7pt} l @{\hskip7pt} l @{\hskip2pt}}
$\alpha$ & $\beta$ & $\epsilon$ \\ 
\hline
0.2 & 0.05 & 3 \\
0.8 & 0.05 & 3 \\
0 & 200 & 1 \\
100 & 1000 & 3 \\
20 & 40 & 3 \\
5 & 20 & 3 \\
1 & 10 & 3 \\
80 & 100 & 5 \\
- & - & - \\
\end{tabular}
&
\begin{tabular}[t]{@{\hskip0pt} l @{\hskip7pt} l @{\hskip7pt} l @{\hskip2pt}}
$\alpha$ & $\beta$ & $\epsilon$ \\ 
\hline
0.4 & 0.2 & 3 \\
0.8 & 0.05 & 3 \\
0 & 200 & 1 \\
100 & 1000 & 3 \\
6 & 12 & 3 \\
9 & 15 & 3 \\
0.1 & 10 & 3 \\
80 & 100 & 5 \\
600 & 170 & 5 \\[0.7ex]
\end{tabular}
\\
Time cost factor & \centering 0.05 & 0.01 & 0.05 \\
  
\bottomrule
\end{tabular}
\end{table}

\subsection{Loitering Mission: An 81-hour solar-powered flight}
\label{sec:Results_81hFlight}

As described in our previous work \cite{Oettershagen_JFR2017} and the corresponding video\footnote{\url{https://www.youtube.com/watch?v=8m4_NpTQn0E}}, \emph{AtlantikSolar} performed its world-record 81-hour continuous solar powered flight in summer 2015 to demonstrate that today's UAV technology allows multi-day solar-powered flights with sufficient energetic safety margins. By demonstrating multi-day station-keeping, the flight set an example for telecommunications relay or aerial observation missions as later on shown in \cite{Oettershagen_JFR2018}. Clearly, such long-endurance missions require careful planning. While the trajectory is fixed, the overall mission feasibility in terms of environmental and energetic safety margins still needs to be assessed and the optimal launch date needs to be found.

\begin{figure}[hp]
\centering
\includegraphics[width=\textwidth]{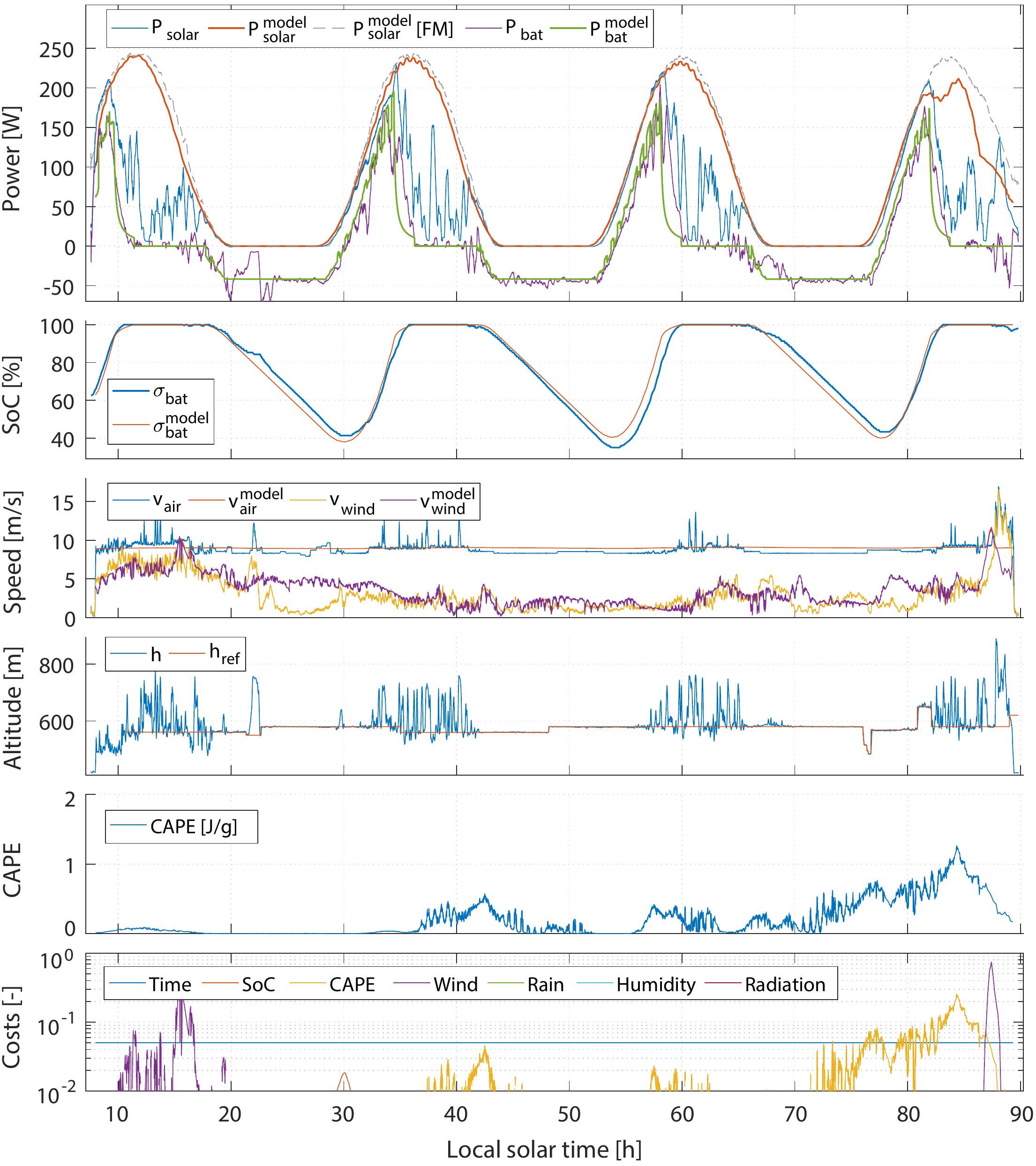}
\caption{Flight data from \emph{AtlantikSolar}'s 81-hour continuous solar-powered loitering mission compared to the MetPASS planning results. The historical weather data used here correctly recovers the clear weather for days one to three and strong winds and thunderstorm clouds for the last day. The top plot shows solar power income $P_\text{solar}$(measured), $P_\text{solar}^\text{model}$(MetPASS), the clear-sky \emph{full model} $P_\text{solar}^\text{model}\text{[FM]}$ from \cite{Oettershagen_2017_SolarModeling}, and battery power. The remaining plots show the state of charge, airspeed, wind conditions, the thunderstorm indicator CAPE and the individual MetPASS costs that especially the strong winds and high CAPE cause. Overall, the MetPASS predictions closely fit the flight data. Used with accurate weather data, MetPASS can thus predict and avoid unsafe situations such as the high winds close to thunderstorms on the last day.}
\label{fig:Results_81hFlight1}
\end{figure}

MetPASS was used to perform both tasks. Using the weather \emph{forecast} available before the flight, MetPASS showed that a 4-day weather window could be leveraged. Launch was performed on June 14th 2015 at 9:32 local time (8.00 solar time) at Rafz, Switzerland. The flight was completed successfully 81.5 hours later on July 17th at 18:58. While the weather forecasts were accurate for the first three days, they did not predict the thunderstorms and severe winds on the last day. To analyze and verify the system models of \cref{sec:Design} using correct weather data, \cref{fig:Results_81hFlight1} thus shows flight data and the MetPASS output based on \emph{historical} weather data (i.e. an a-posteriori weather analysis of the COSMO-2 model with \unit[2]{km} spatial and \unit[1]{h} time resolution).

Given the clear-sky conditions MetPASS predicts a solar power income $P_\text{solar}^\text{model}$ close to the theoretical maximum represented by the full solar power model $P_\text{solar}^\text{model}\text{[FM]}$ developed in \cite{Oettershagen_2017_SolarModeling}. The decrease of $P_\text{solar}^\text{model}$ due to clouds on the last day is captured correctly by MetPASS. The measured $P_\text{solar}$ also closely follows the MetPASS predictions. It only deviates when $\text{SoC}\approx\unit[100]{\%}$ because $P_\text{solar}$ is throttled down as per design to protect the batteries. Solar power is therefore only supplied to cover the fluctuating propulsion demands. The required battery power to sustain flight during the night is $P_\text{bat}=\unit[41.6]{W}$ whereas MetPASS estimates $P_\text{bat}^\text{model}=\unit[42.4]{W}$ using the measured power-curve fitted to \cref{eqn:P_level}. The overall charge and discharge process is thus represented very accurately. The predicted and measured minimum SoCs averaged over all three nights are \unit[39]{\%} and \unit[40]{\%} respectively. Notable deviations are only caused by unexpected evening thermals, which are visible through altitude fluctuations and a decrease in $P_\text{bat}$, during the first and third night. Overall, the MetPASS predictions represent the measurements very well. Note however that given the mostly excellent weather conditions, these energetic results would not even have required taking weather forecasts into account.

In contrast, the wind and airspeed plots show more variety due to the current weather. Strong winds are measured and modeled for the first day. In accordance with the ground operators on the field, at $t=\unit[15.4]{h}$ solar time MetPASS suggests to increase the airspeed to assure $v_\text{gnd}>0$. The winds decrease for the following nights, but then reach up to \unitfrac[11.4]{m}{s} (model) and \unitfrac[16]{m}{s} (measurement) during the last two flight hours. With a nominal airspeed of only $v_\text{air}\approx\unitfrac[9]{m}{s}$, both MetPASS and the ground operators increase $v_\text{air}$ to keep $v_\text{air}\geq v_\text{wind}$ to avoid the vehicle's drift-off. In contrast to MetPASS, the ground operators increase the UAV's altitude to later convert it into more speed if the wind picks up even more. The higher altitude is one reason why the UAV measures higher wind speeds. In addition, as clearly indicated by the convective available potential energy (CAPE), thunderstorm clouds develop. Overall, the costs for wind and thunderstorms are close to their normalized limit ($\dot{C}_k=1$). The costs correctly represent the significant danger the UAV is in, both because of a possible drift-off and a structural overload due to the strong wind shear and gusts. As mentioned before, these environmental conditions were unfortunately not predicted by the initial weather forecast from four days earlier and the ground crew did not exploit updated weather data. The main lesson learned from this flight is thus that --- especially for multi-day missions during which the weather can change significantly --- MetPASS's re-planning capability is \emph{key} and \emph{needs} to be used to avoid all unsafe situations for solar aircraft. In this case, the UAV could have simply been landed before the high winds and thunderstorm clouds arrived.
 
\subsection{Point-to-Point Mission: Crossing the Atlantic Ocean}
\label{sec:Results_Atlantic}

Crossing the Atlantic Ocean is a feat that not many small-scale UAVs can accomplish. It is therefore an excellent demonstration case for long-endurance solar UAVs such as \emph{AtlantikSolar}\footnote{The Atlantic crossing inspired the name \emph{AtlantikSolar} UAV. While found to be technically feasible, the mission was not executed because of regulatory reasons. More specifically, the \unit[4000]{km} flight in BVLOS conditions would not have received regulatory approval from the transport authorities because of the requirement for sense-and-avoid capabilities on par with humans, which is technically not feasible on today's UAVs.}. The \unit[4000]{km} route from Newfoundland, Canada to Lisbon, Portugal was selected as \emph{AtlantikSolar}'s prime --- and \emph{design-driving} --- mission. Clearly, such a flight involves significant challenges. In addition to careful regulatory planning, extensive ground infrastructure and an efficient yet robust system design, a thorough pre-assessment of the exact times and conditions under which the flight is feasible is required. The fragility and weather-susceptibility of solar UAVs also demands that the weather is carefully monitored along the whole route and new trajectories that avoid upcoming severe weather are generated in flight. MetPASS was used for both these tasks.

\paragraph{Historical weather data: Determining optimal and marginal flight conditions}

\begin{figure}
\centering
\includegraphics[width=\textwidth]{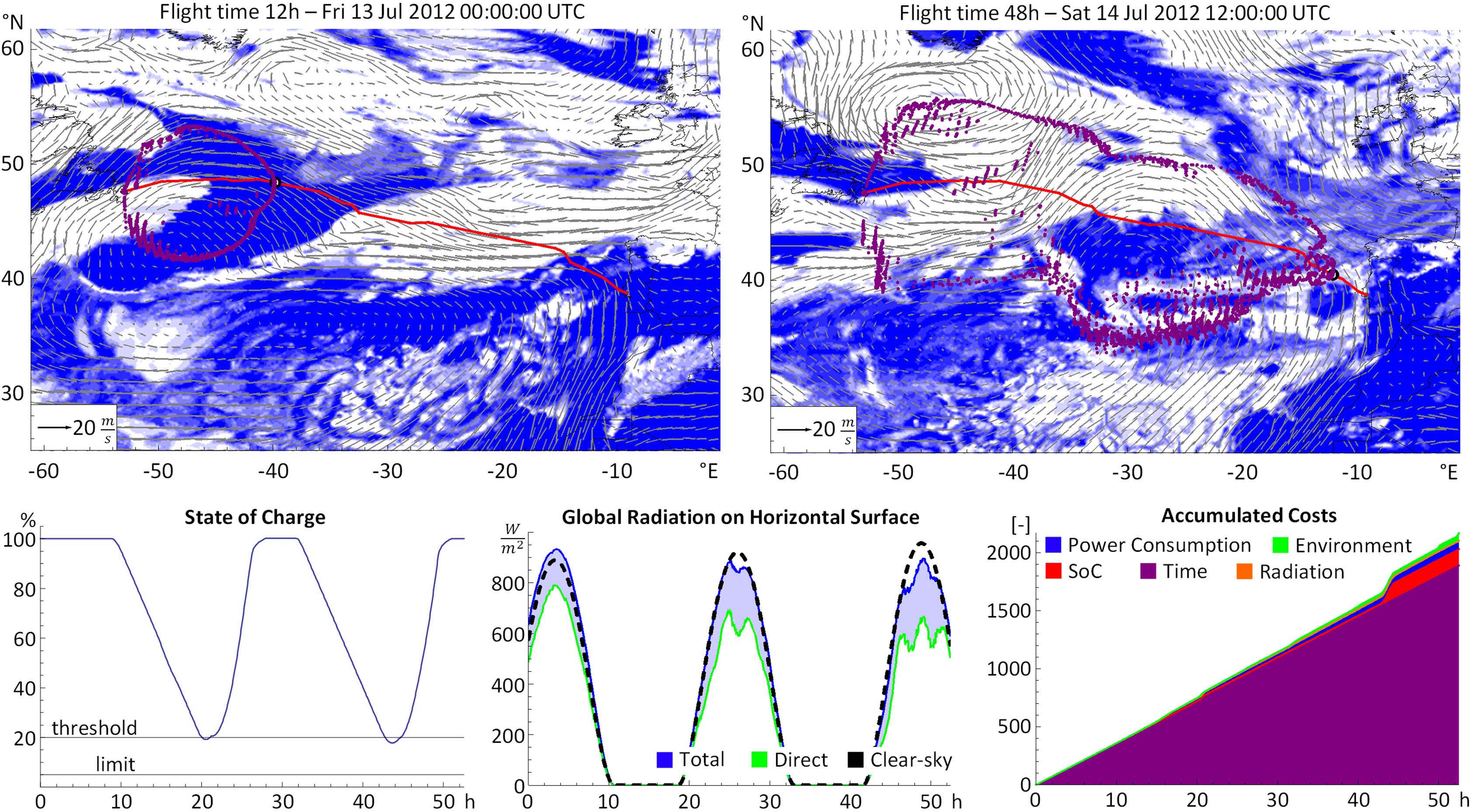}
\caption{A transatlantic flight under optimal weather conditions: MetPASS exploits the easterly winds and thereby retrieves an orthodrome-like path with only 52 hours flight time. The \ac{SoC} barely falls below the threshold. Except for the time cost, all costs are small. This proves that the Atlantic crossing is feasible even with the \emph{AtlantikSolar} AS-1 UAV if the right launch time is chosen with the help of the planner.}
\label{fig:Results_Atlantic1}
\end{figure}

\begin{figure}
\centering
\includegraphics[width=\textwidth]{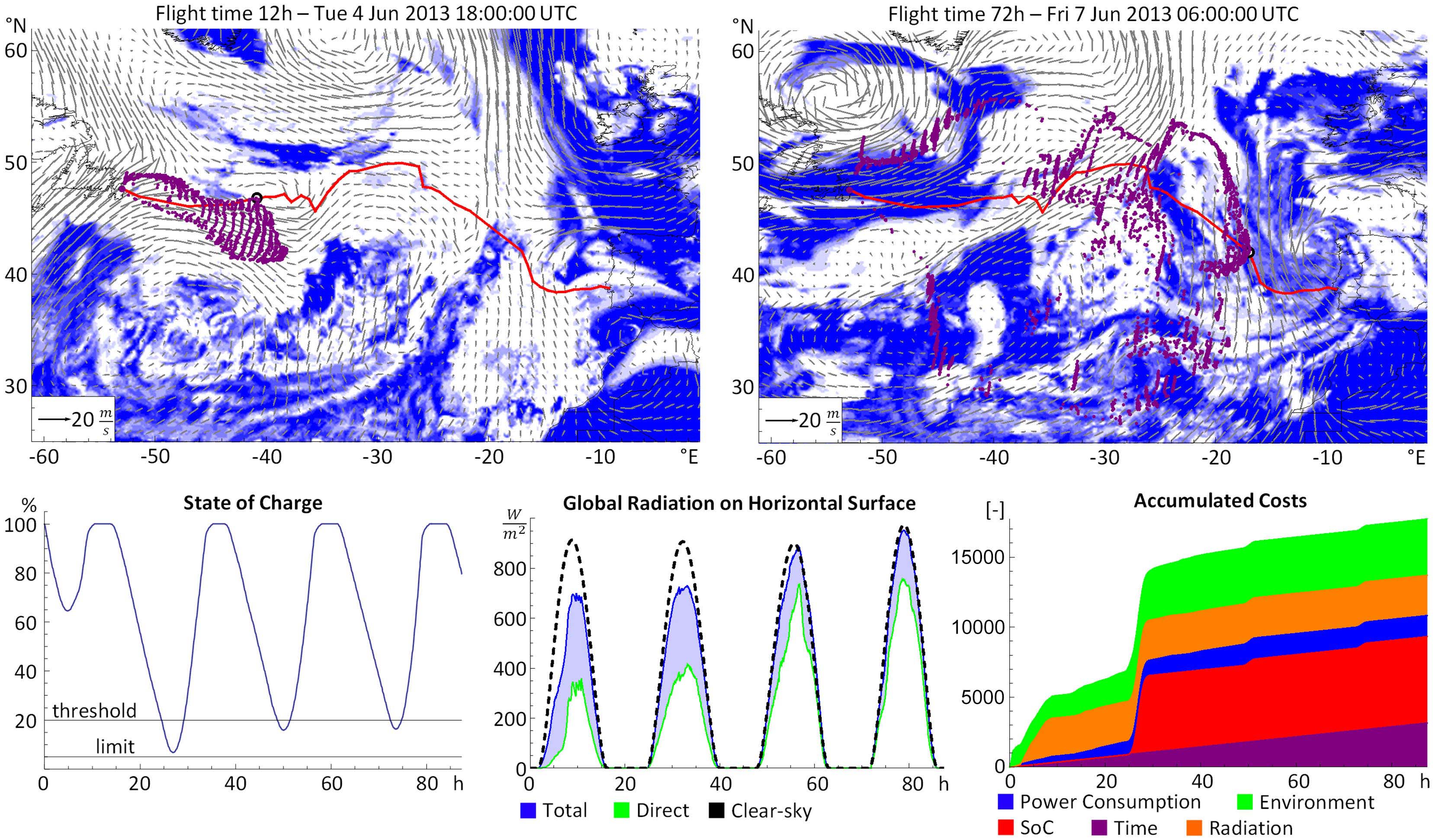}
\caption{A transatlantic flight under unsuitable weather conditions: The launch happens in heavy winds and under high cloud coverage. In the first night, the \ac{SoC} is critically low. The total costs for \ac{SoC}, power consumption, radiation and environmental dangers are an order of magnitude higher than for the optimal weather case. A feasible path is found, but the result clearly encourages to choose a different launch date using MetPASS.
}
\label{fig:Results_Atlantic2}
\end{figure}

To identify the full range of conditions under which an Atlantic crossing is feasible, MetPASS was applied to historical ECMWF weather data from 2012 and 2013. Optimal and marginal border-cases together with their performance metrics minimum \ac{SoC}, total accumulated cost and required flight time were identified. \cref{fig:Results_Atlantic1} shows an exemplary optimal border case on July 13th, 2012: The chosen route closely follows the orthodrome, with tailwind reducing the flight time by more than \unit[50]{\%} to \unit[52]{hours}. Because the planner chose the launch time correctly, significant cloud cover can be avoided and $P_\text{solar}\approx P_\text{solar}^\text{clear-sky}$. The state of charge therefore always stays above \unit[17.2]{\%}. The accumulated cost is $C=2200$ and mainly consists of the time cost, which indicates that all other costs usually stay below their threshold. In contrast, the exemplary June 4th, 2013 test case in \cref{fig:Results_Atlantic2} exhibits marginal conditions. To avoid unsuitable areas (severe cross and headwind, high humidity and low solar radiation), the planner chooses a path that deviates significantly from the orthodrome. The resulting flight time is \unit[86]{h}. The minimum state of charge is only \unit[7.1]{\%}. Although none of the costs reaches the critical limit $\beta_k$, the total accumulated cost are $C=17800$. Launch is not recommended under these conditions. As described below, a more optimal nearby  launch date can easily be determined using MetPASS.

\paragraph{Historical weather data: Determining the seasonal dependency of feasibility}

\begin{figure}[htb]
\centering
\includegraphics[width=\textwidth]{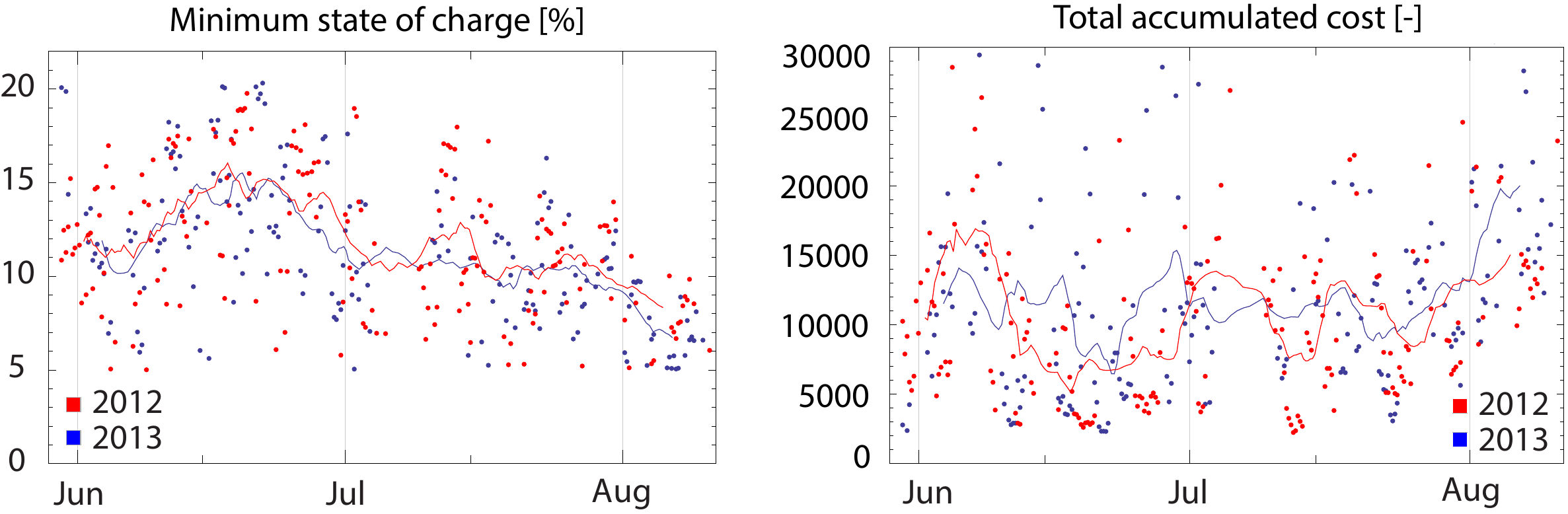}
\caption{
Feasibility assessment for the Atlantic crossing over the summer seasons of 2012--2013. Minimum \ac{SoC} (left) and total accumulated cost (right). Solid lines indicate the average of 20 surrounding days. As expected, feasibility is highest around June 21st.
}
\label{fig:Results_Atlantic3}
\end{figure}

Due to the variety of weather conditions above the Atlantic it has to be assessed systematically how many and when launch windows for such a flight exist. The seasonal dependency of the performance metrics was thus assessed via MetPASS trajectory optimizations that were run in \unit[6]{h} steps for the whole range from May 31st to August 8th for which historical ECMWF data was available. \Cref{fig:Results_Atlantic3} shows the minimum \ac{SoC} and total accumulated cost for each date. It can be inferred that even the first version of the \emph{AtlantikSolar} UAV (AS-1, see \cite{Oettershagen_ICRA2015}) provides sufficient feasible launch dates from mid-May to end-July when requiring a state of charge margin of \unit[10]{\%}. Obviously, the later versions of \emph{AtlantikSolar} (AS-2 and AS-3, see \cite{Oettershagen_JFR2017} and \cite{Oettershagen_JFR2018}) would improve the performance metrics. Additional analysis yields a minimum and average flight time of \unit[52]{h} and \unit[78]{h} respectively versus the \unit[106]{h} for the no-wind unit test of \cref{sec:Impl_Validation}. \Cref{fig:Results_Atlantic3} can obviously also be used before a flight to decide how optimal a certain launch date is relative to all other launch dates of the season.

\paragraph{Using forecasts: Launch-time optimization and real-time route re-planning}

\begin{figure}[htb]
\centering
\includegraphics[width=\textwidth]{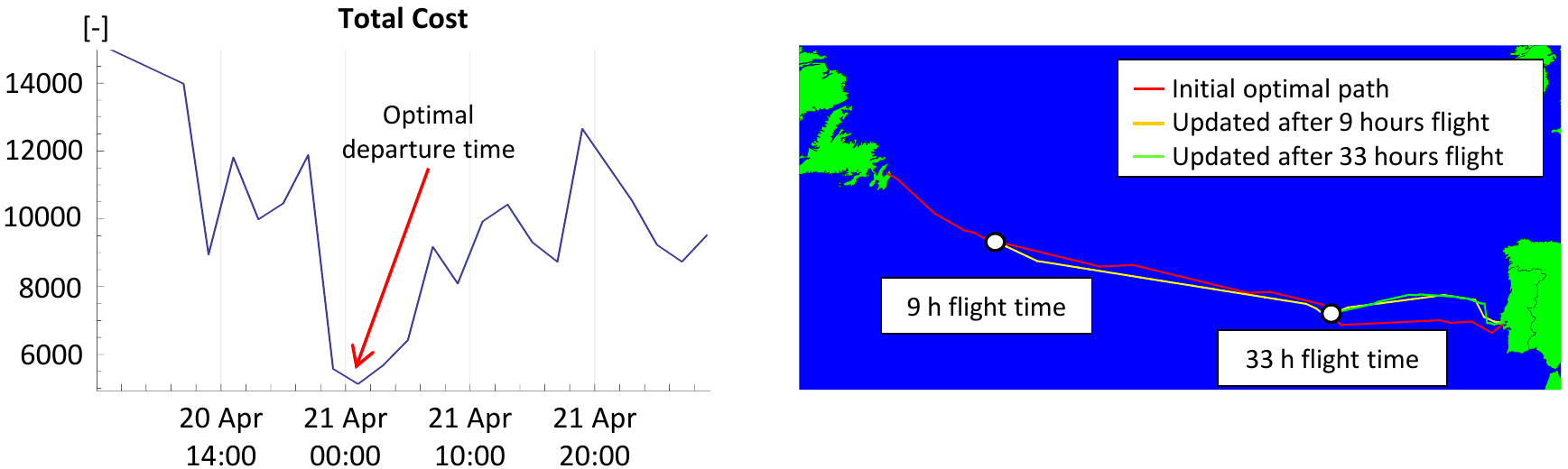}
\caption{
MetPASS end-to-end mission planning process: The optimal departure time is determined based on current weather forecasts (left). Once the UAV is airborne, the route is replanned periodically using updated weather forecasts (right) and the waypoints are re-sent to the vehicle. 
}
\label{fig:Results_Atlantic4}
\end{figure}

\Cref{fig:Results_Atlantic4} shows MetPASS's ability to perform launch time optimizations as well as in-flight route-corrections using periodically-updated forecast data. For the success of a mission, the optimal launch time is as important as the path itself. In \cref{fig:Results_Atlantic4}, the total cost was therefore calculated for a 50 hour time window around April 21st, 2014. The time with minimum total cost is chosen as the launch time and the corresponding path can then be checked and executed by the aircraft. Given the significant weather fluctuations during long-endurance missions, 9 hours after launch the path optimization is restarted with updated weather data but the current aircraft state (retrieved via telemetry) as initial values. This path correction, also illustrated in \cref{fig:Results_Atlantic4}, is repeated whenever new weather data arrives (every 24 hours in this specific case) until the destination is reached.

\subsection{Multi-Goal Missions: Inspecting Arctic Glaciers}
\label{sec:Results_Arctic}

Large-scale missions that involve scanning multiple areas of interest can optimally leverage the long-endurance capabilities of solar aircraft. For example, the persistent aerial monitoring of Arctic glaciers and the analysis of their flow and calving characteristics is key to understanding global climate change. Previously, glaciologists needed to use expensive on-site transport (e.g. helicopters) and needed to operate in remote places with limited infrastructure (tents without internet) next to the glaciers \cite{Jouvet_BowdoinGlacier2015}. The central goal for \emph{AtlantikSolar}'s deployment to the Arctic Ocean in summer 2017 was thus to demonstrate a new paradigm for glacier research: The operation of a complete scanning mission, i.e. take-off, inspection of one or even more remote glaciers, and landing, from an easily-accessible home base. The home base was chosen as Qaanaaq, a village at a latitude of \unit[77]{\degree N} in Northwest-Greenland and surrounded by multiple fast-flowing Arctic glaciers (\cref{fig:Intro_Collage}). A number of scanning missions with increasing complexity were performed, two of which are described below. MetPASS performed both the initial feasibility assessment as well as the launch date and route optimization over the Arctic Ocean.


\subsubsection{Two-glacier mission: Bowdoin Glacier}
\label{sec:Results_Arctic_Bowdoin}

The inspection mission from Qaanaaq to Bowdoin glacier was performed on July 3rd 2017. After days of fog and excessively strong Arctic winds, clear weather with only little high altitude clouds followed. Using weather forecasts from \unit[15]{hours} before takeoff, a launch window was found, a path was generated and take-off was performed at 16:26 local time (13.76h solar time). The mission objective was to perform two different scans: One higher altitude lawn-mower scan to observe possible glacier calving events at Bowdoin glacier, and one low-altitude station-keeping scan at base. 

\begin{figure}[htb]
\centering
\includegraphics[width=0.75\textwidth]{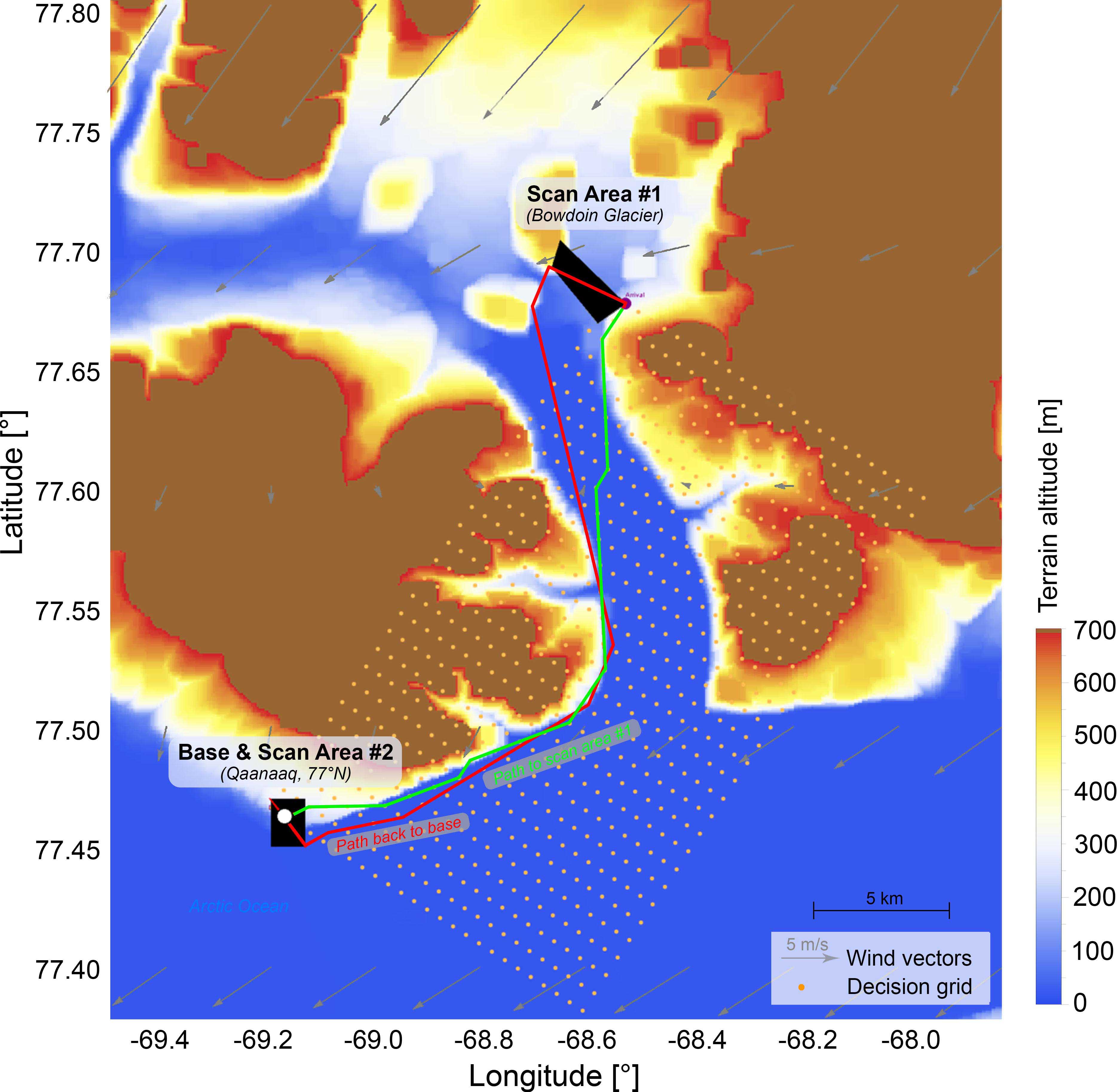}
\caption{The \emph{AtlantikSolar} glacier inspection mission above the Arctic Ocean planned by MetPASS. The small scale of the problem and the narrow fjords clearly reduce the path choices. The result is thus a compromise that yields the shortest path and avoids terrain. 
}
\label{fig:Results_ArcticBowdoinPath}
\end{figure}

The corresponding MetPASS path planned with the parameters of \cref{tab:Results_OverviewAndParameters} (i.e. 30 slices times 25 vertices per grid) is shown in \cref{fig:Results_ArcticBowdoinPath}. In contrast to our previous work~\cite{Wirth_AeroConf2015}, the planner was extended with the ability to avoid the terrain represented by a \unit[30]{m} resolution \acl{DEM}. In addition, while by default the departure and arrival points are initialized at the center of the first and last slice of the grid (see \cref{fig:Design_Grid}), this can lead to large amounts of grid points over inaccessible terrain. Automatic grid shifting, which maximizes the amount of grid points over accessible areas such as the open ocean, was thus implemented. The optimal path found is a shortest-time path that stays safe of terrain. In other words, the cost function components \emph{time} and \emph{terrain} dominate the path optimization. This is due to, first, the narrow fjord-like terrain which severely limits the path choices when assuming a fixed altitude. Second, due to the relatively small scale of the planning problem (compared e.g. to \cref{sec:Results_Atlantic}) the weather is rather homogeneous over the whole area and exploiting these small weather differences yields less cost advantages than following a time-optimal path.

\begin{figure}[htbp]
\centering
\includegraphics[width=\textwidth]{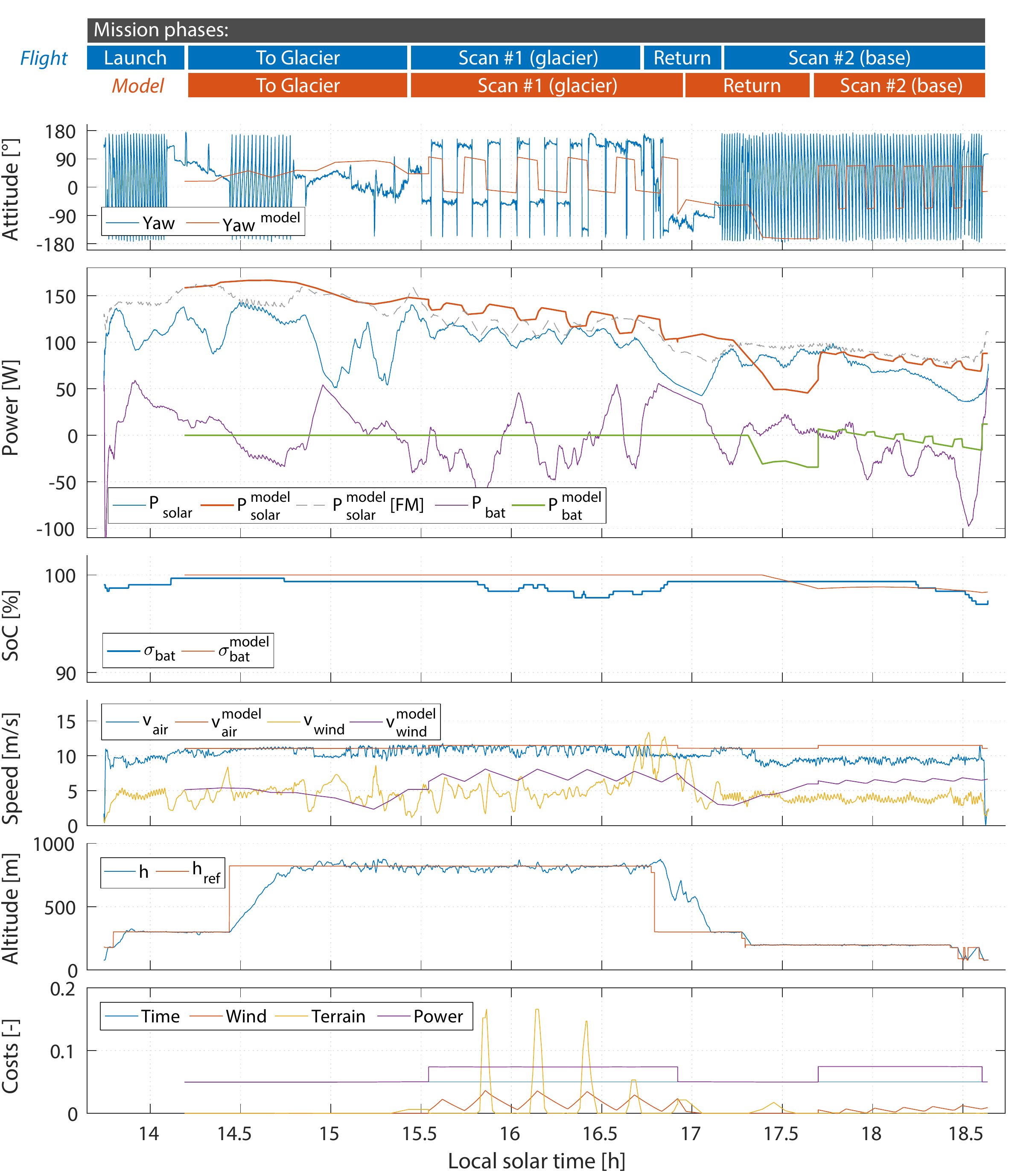}
\caption{The \unit[4:52]{h} and \unit[230]{km} glacier inspection mission performed by \emph{AtlantikSolar} on July 3rd 2017 above the Arctic Ocean. The flight phases are indicated in blue for the flight data and in red for the MetPASS plan. They include start at the base in Qaanaaq (Greenland), traveling to and scanning Bowdoin glacier and returning to base to perform a second aerial scan before landing. The MetPASS plan uses weather forecasts from \unit[15]{hours} before launch. The energetic states are represented well, but erroneous wind forecasts cause significant differences in the time required to complete each individual flight phase. To preserve comparability, the MetPASS plan was therefore adapted to the same duration as the actual flight (only through shortening the second scanning phase a-posteriori).
}
\label{fig:Results_ArcticBowdoin}
\end{figure}

\Cref{fig:Results_ArcticBowdoin} compares the flight data recorded by \emph{AtlantikSolar} against the MetPASS predictions. On the one hand, solar and battery power $P_\text{solar}$ and $P_\text{bat}$ are again represented accurately. We measure $P_\text{solar}<P_\text{solar}^\text{model}$ for $t=[\unit[13.8]{h},\unit[15.6]{h}]$ because $SoC\approx\unit[100]{\%}$ and $P_\text{solar}$ is thus again throttled down by design. The battery discharge starts around \unit[18.0]{h} solar time, where $P_\text{bat}<P_\text{bat}^\text{model}$ because of upcoming high altitude clouds and the setting sun. On the other hand, the wind forecast is subject to significant errors: At $t\approx\unit[16.8]{h}$, the measured and predicted winds are $v_\text{wind}=\unitfrac[13.3]{m}{s}$ and $v_\text{wind}^\text{model}=\unitfrac[8.0]{m}{s}$ (at the northern scan path section at Bowdoin, \cref{fig:Results_ArcticBowdoinPath}). The time required to complete each flight phase thus differs between the model and flight data. More importantly, wind speeds of this magnitude are a significant threat to the aircraft. While the predicted maximum winds of $v_\text{wind}^\text{model}=\unitfrac[8.0]{m}{s}$ had already been accounted for by flying at $v_\text{air}\approx\unitfrac[11]{m}{s}>v_\text{air}^\text{opt}$, the flight would have been marked as infeasible by MetPASS had the real wind speeds been known. Again, the higher wind speeds would have been predicted by the forecasts available \unit[3]{h} before launch. A lesson learned from this flight is therefore that, again, the most up-to-date weather data always needs to be used. More generally speaking, only high quality weather data brings tangible benefits when planning such missions. 

\begin{figure}[htb]
\centering
\subfloat[]{\includegraphics[width=0.49\textwidth]{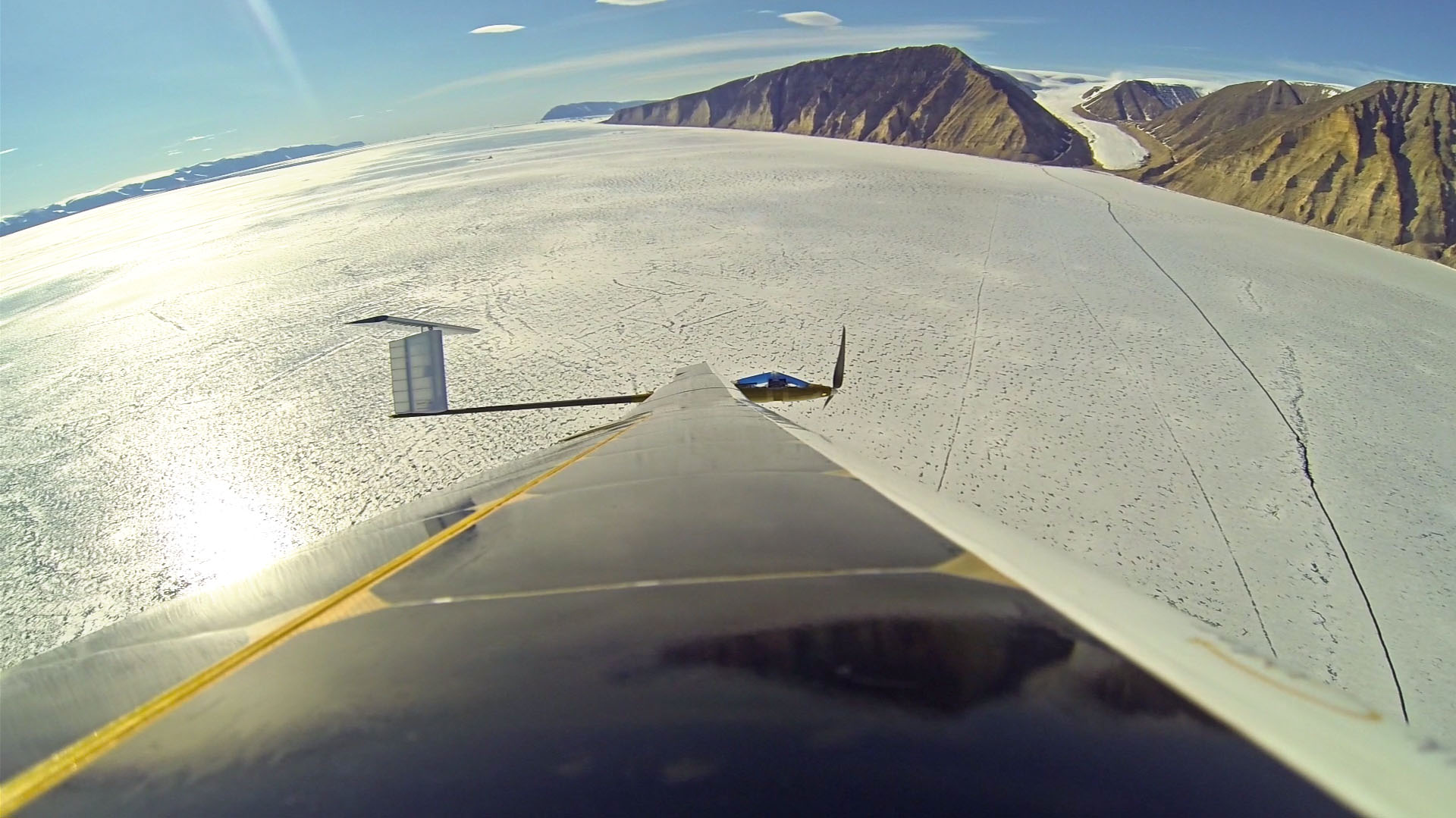}}
\hspace*{\fill} 
\subfloat[]{\includegraphics[width=0.49\textwidth]{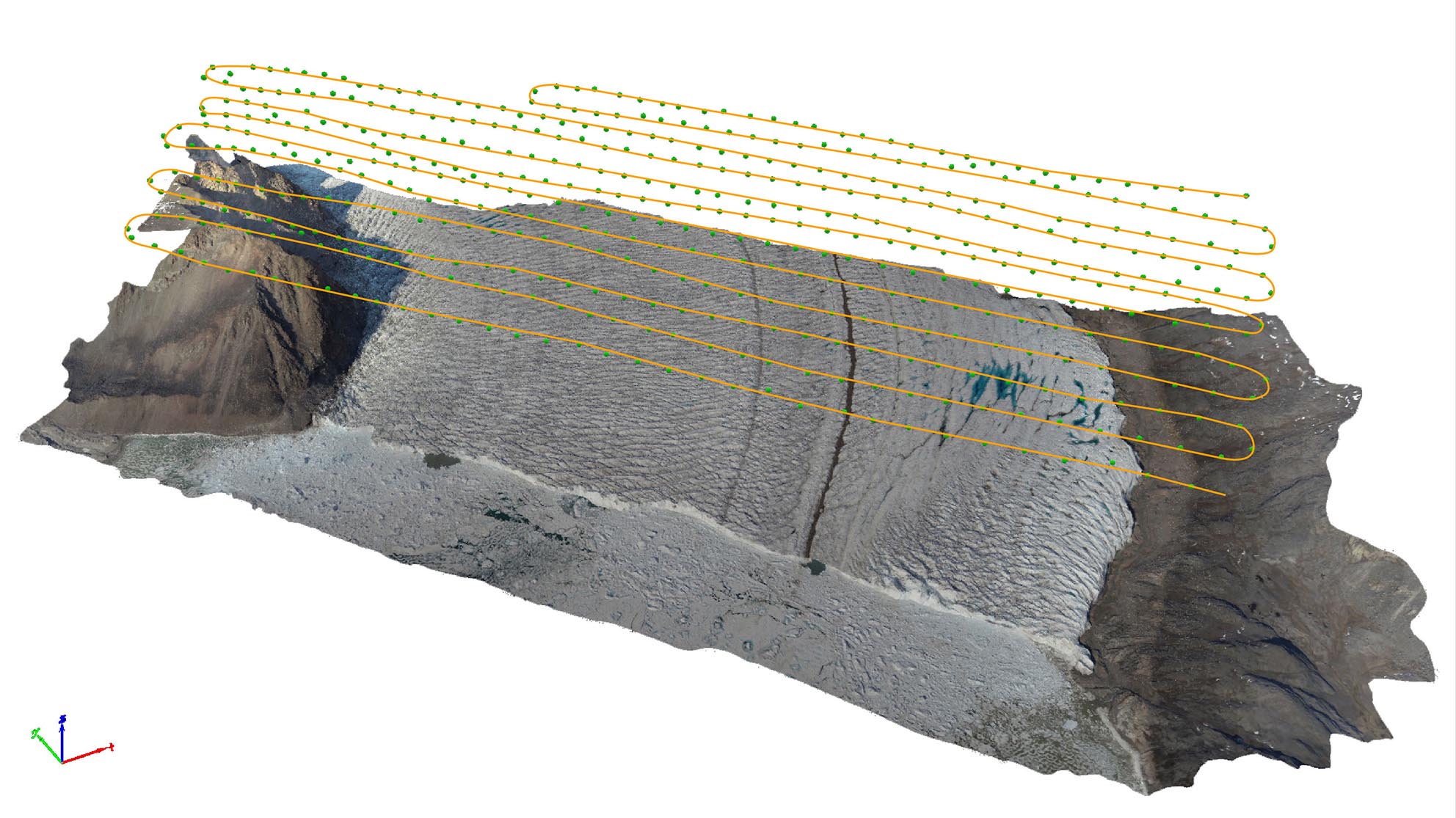}\label{fig:Results_ArcticBowdoinPictures2}}
\caption{Left: The \emph{AtlantikSolar} UAV operating over the Arctic sea ice next to Bowdoin glacier. Right: The full 3D construction (made with the commercially available Pix4D software) of Bowdoin glacier and the scan path planned by MetPASS. The scan area of $\unit[7]{km^2}$ was covered with \unit[75]{\%} lateral overlap from \unit[700]{m} AGL, which took 53 minutes and covered \unit[47.7]{km}.
}
\label{fig:Results_ArcticBowdoinPictures}
\end{figure}

Overall, using the MetPASS flight plan, this first-ever glacier inspection mission with a solar-powered UAV in the Arctic could be executed successfully (\cref{fig:Results_ArcticBowdoinPictures}). As predicted, the main costs or threats for the airplane were the high wind speeds at $t=[\unit[15.5]{h},\unit[17.0]{h}]$ and the high terrain ($t=[\unit[15.8]{h},\unit[16.8]{h}]$), which the MetPASS path avoided successfully were possible. Other environmental costs were not significant. The overall flight duration was \unit[4:52]{h} during which \unit[230]{km} were covered. Even after that, the batteries were still almost full ($\text{SoC}=\unit[97]{\%}$). The whole mission was performed fully autonomously, i.e. without any pilot intervention. The scientific results provided to glaciologists were a full 3D reconstruction of Bowdoin glacier (\cref{fig:Results_ArcticBowdoinPictures2}) which showed a developing crack that led to a significant glacier calving event only days after the flight. Overall, the mission objective of demonstrating a fully-autonomous end-to-end aerial scanning mission of remote glaciers from a local home base could thus be fulfilled successfully.

\subsubsection{Six-glacier mission}
\label{sec:Results_Arctic_TheHeilprins}

MetPASS was also leveraged to plan a multi-goal mission involving the aerial scanning of the six Arctic glaciers \code{A--F} in \cref{fig:Intro_Collage}. The settings in \cref{tab:Results_OverviewAndParameters} were employed, i.e. the point-to-point grid was reduced to 12 slices and 9 vertices to guarantee low calculation times. The mission was started at 2:30 local time (4:30 UTC) on July 6th 2017. Two glaciers (Heilprin North and Heilprin South) were scanned successfully, but due to state estimation issues that occurred after 7 hours of flight the scanning of further glaciers had to be canceled. Although the mission could not be completed successfully, the topology of such a solar-powered multi-glacier scanning mission in the Arctic is of course still instructive and is thus described below.

\begin{figure}[htbp]
\centering
\includegraphics[width=0.98\textwidth]{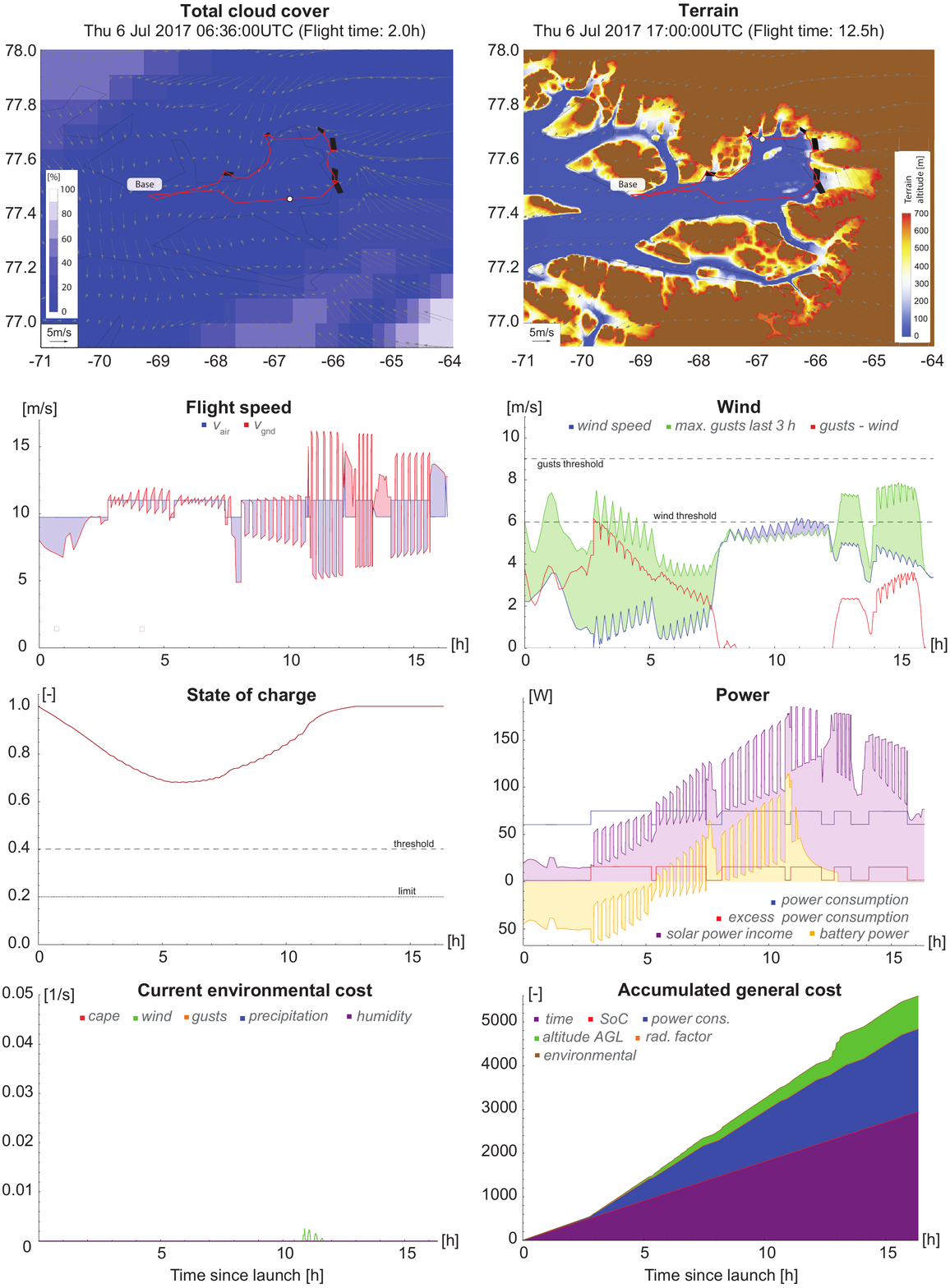}
\caption{The six-glacier scan mission in the Arctic optimized with MetPASS. The path is shown against cloud cover (top left) and terrain (top right). The weather conditions are favorable: The environmental costs are close to zero and the main costs are excess power consumption (due to winds), altitude above ground and time. Despite strong winds the \ac{SoC} never dips below \unit[68]{\%}, indicating the potential of solar-powered flight in Arctic regions.
}
\label{fig:Results_ArcticMultiGlacier}
\end{figure}

\Cref{fig:Results_ArcticMultiGlacier} shows the optimal path calculated by MetPASS. The total path covers \unit[580]{km} distance in \unit[16.3]{h} flight time. The aircraft position is shown both at 6:36 UTC and 17:00 UTC, once overlaid with a total cloud cover map and once with a terrain map. The cloud cover and thus sun radiation situation is very favorable, however, significant winds are indicated next to the areas of interest. The operators therefore increase the flight speed from the optimal $v_\text{air}^\text{opt}=\unitfrac[8.6]{m}{s}$ to $v_\text{air}=\unitfrac[9.8]{m}{s}$ while traveling and $v_\text{air}=\unitfrac[11]{m}{s}$ inside the areas of interest. The optimal path found by MetPASS is near counter-clockwise and thus clearly exploits the fact that the meso-scale winds exhibit a counter-clockwise rotation. In addition, when the straight-line path between two areas of interest would result in headwind, the aircraft also deviates from the straight line path (i.e. at around t=6:36 UTC, see \cref{fig:Results_ArcticMultiGlacier} top left, a path south of the straight line path allows to avoid headwinds). The flight speed and wind plots  therefore show that the aircraft is progressing fast. More importantly, the path manages to completely avoid dangerous high-wind areas (except at t=11.2h after launch). The power income is high but is, as before, heavily influenced by the aircraft heading due to the low sun elevation in the Arctic. The state of charge never drops below \unit[68]{\%} despite launching around solar midnight, thus confirming the potential of solar-powered flight in Arctic regions. The total accumulated cost is $C=5595$. The largest costs are time, excess power consumption (caused when $v_\text{air}>v_\text{air}^\text{opt}$), and altitude AGL (due to flight above high terrain next to the glaciers). 
Overall, the combination of good weather and the path optimized by MetPASS allows to avoid environmental risks and only results in \emph{acceptable} costs that either increase flight safety (increased airspeed and thus power consumption) or cannot be avoided (altitude AGL next to the glaciers and time). 

To assess the advantages of this optimal path versus more naive solutions, \cref{tab:Results_ArcticTable} compares the corresponding path metrics cost, distance and time. As expected, the MetPASS path is cost-optimal. Although it is very similar to the counter-clockwise path, the fact that it starts visiting the nodes in the order \code{EF\hspace{-0.3ex}.\hspace{-0.5ex}.\hspace{-0.5ex}.} instead of \code{FE\hspace{-0.0ex}.\hspace{-0.5ex}.\hspace{-0.5ex}.} allows it to better avoid certain bad weather phenomena (in this case strong wind) and altitude AGL costs. However, the total cost difference to the naive solutions is only one percent. The reason is that the weather conditions are favorable and no critical environmental risks (e.g. thunderstorms or precipitation) exist that would be avoided by MetPASS but that would incur high costs in a naive path. The overall distance flown is also similar due to the similarity between the paths. In comparison to the clockwise path, which always commands the aircraft to fly against the counter-clockwise-rotating global winds, the MetPASS path however features a \unit[5]{\%} shorter overall flight time. Note that in such good weather conditions, other less trivial paths (i.e. those not strictly clockwise or counter-clockwise) always result in additional flight time and thus cost and therefore cannot be optimal. 

\begin{table}[htb] 
\caption{The optimal path found by MetPASS against the naive solutions.} \label{tab:Results_ArcticTable}
\centering
\begin{tabular}{l l l l}
\toprule
& MetPASS & Clockwise & Counter-clockwise \\ 
\midrule
Cost & 5595 & 5634 (\unit[+0.7]{\%}) & 5650 (\unit[+1.0]{\%})\\
Time & \unit[16.3]{h} & \unit[17.1]{h} (\unit[+4.9]{\%})& \unit[16.2]{h} (\unit[-0.6]{\%})\\
Distance & \unit[580]{km} & \unit[579]{km} (\unit[-0.2]{\%}) & \unit[572]{km} (\unit[-1.4]{\%})\\
Order & EFDCBA & ABCDEF & FEDCBA \\
\bottomrule
\end{tabular}
\end{table}

\subsection{Computational Performance Analysis}
\label{sec:Results_PerformanceAnalysis}

The methods and framework developed in this paper shall be usable for pre-flight mission planning as well as in-flight re-planning. The required computation time, which is shown in \cref{tab:Results_ComputationTimes} for the missions presented before, therefore plays a crucial role. First, station-keeping missions do not require actual route planning but only straightforward system state propagation. Even multi-day station-keeping missions such as the 81h-flight can therefore be calculated in a couple of seconds. Second, large-scale point-to-point missions such as the \unit[4000]{km} trajectory across the Atlantic Ocean naturally require point-to-point route optimization. Under the parameters of \cref{tab:Results_OverviewAndParameters}, MetPASS completes the route optimization in less than 10 minutes. Third, the multi-goal aerial inspection missions require the scan path optimization, heuristics calculation, and an inter-goal optimization that includes either $n_\text{heuristic}$ or $n_\text{naive}$ point-to-point route optimizations depending on whether the heuristic is used or not. The two-glacier scan mission is optimized within 11 minutes. The heuristic does not provide a computational advantage for this case because a) it takes much longer to calculate the heuristic than all point-to-point optimizations and b) the heuristic does not even reduce the number of edge cost evaluations. For the six-glacier mission, the situation changes drastically: The 9 minutes invested to calculate the heuristic allow to reduce the required edge cost calculations by \unit[94]{\%}. The total calculation time $t_\text{total}$ with heuristic is 17 minutes\footnote{Note however that, compared to the two-glacier mission, the grid resolution was reduced in return for a fast computation.} and is thus, based on a simple extrapolation, 8 times shorter than if all edge cost evaluations had to be performed.  		
 		
\begin{table}[htb] 
\caption{Computation times with a \unit[2.8]{GHz} quad-core Intel Xeon E3-1505M CPU with 16GB RAM and the parameters from \cref{tab:Results_OverviewAndParameters}. The total computation time $t_\text{total}$ consists of the scan path and heuristic computation times $t_\text{scanpaths}$ and $t_\text{heuristic}$ and the route optimization time $t_\text{opt}$ (which includes the inter-goal and point-to-point optimization). The six-glacier mission benefits significantly from using the heuristic.} \label{tab:Results_ComputationTimes}
\centering
\begin{tabular}{l l l l l}
\toprule
 \multirow{2}{*}{\textbf{Mission}} & \textbf{81h-Flight} & \textbf{Atlantic} & \textbf{Arctic} & \textbf{Arctic} \\
 & & & Two glaciers & Six glaciers \\
 \midrule
$t_\text{total}$, of which & \unit[16]{s} & \unit[462]{s} & \unit[690]{s} & \unit[1010]{s} \\
\tiny\textbullet \normalsize \;$t_\text{scanpaths}$ & - & - & \unit[1]{s} & \unit[3]{s} \\
\tiny\textbullet \normalsize \;$t_\text{heuristic}$ & - & - & \unit[579]{s} & \unit[538]{s} \\
\tiny\textbullet \normalsize \;$t_\text{opt}$ & \unit[16]{s} & \unit[462]{s} & \unit[110]{s} & \unit[469]{s} \\
$n_\text{heuristic}$ & - & - & 6 & 161 \\
$n_\text{naive}$ & - & - & 6 & 2676 \\
$\nicefrac{n_\text{heuristic}}{n_\text{naive}}$ & - & - & 1.0 & 0.06 \\
\bottomrule
\end{tabular}
\end{table}

\begin{figure}[htb]
\centering
\subfloat[The heuristic quality distribution computed through 600 experiments with random boundary conditions (departure and arrival coordinates as well as start time). The mean heuristic quality is 20 percent.]{\includegraphics[width=0.48\linewidth]{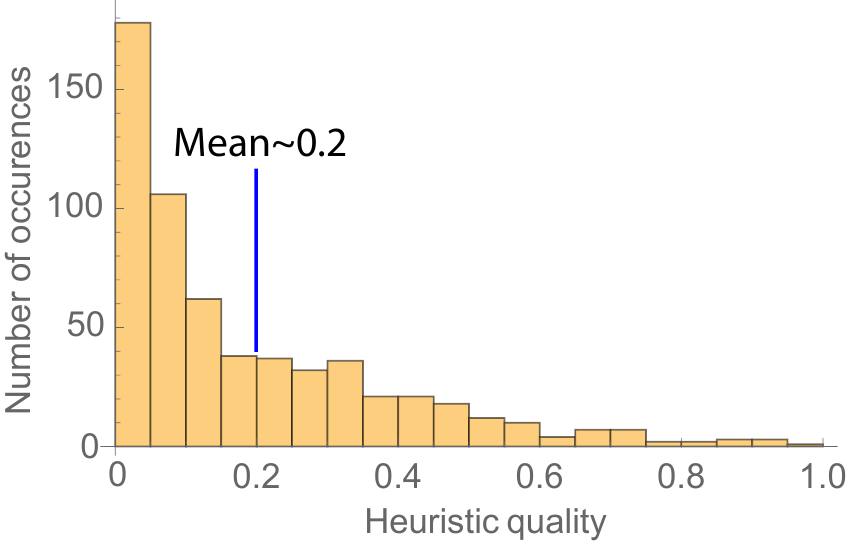}\label{fig:Results_PerformanceAnalysisHeuristic1}}
\hspace*{\fill} 
\subfloat[The number of required edge cost evaluations as a function of heuristic quality and areas of interest $N$. Each data point was obtained by simulating 10 randomly generated problems.]{\includegraphics[width=0.48\linewidth]{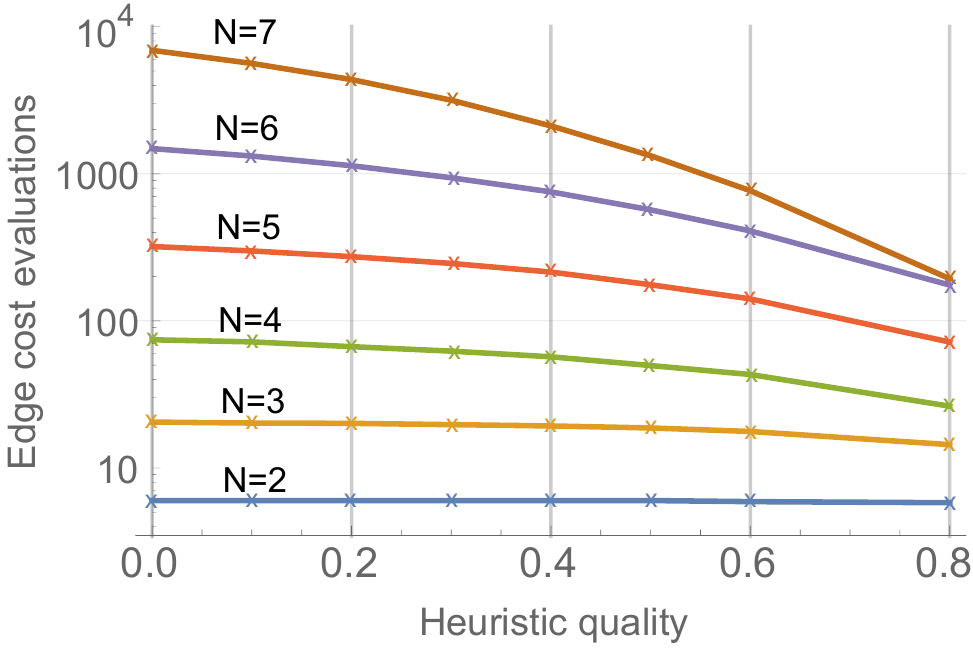}\label{fig:Results_PerformanceAnalysisHeuristic2}}
\caption{Heuristic quality and resulting performance gains through a reduction in edge cost evaluations in MetPASS.
}
\label{fig:Results_PerformanceAnalysisHeuristic}
\end{figure}

\Cref{fig:Results_PerformanceAnalysisHeuristic1} analyzes the effect of the heuristic in more detail. To assess how accurate our heuristic is we define the heuristic quality between vertices $v$ and $w$ as
\begin{equation}
q_\text{heuristic}=\frac{h_{vw}}{c_{vw}}=\frac{\text{heuristic value}}{\text{actual cost}}\; .
\end{equation} 
Here, $q_\text{heuristic}<1$ must hold for $h_{vw}$ to be a valid heuristic, but the closer $q_\text{heuristic}$ gets to one the earlier suboptimal paths can be sorted out. The quality of the heuristic implemented in this work was determined through 600 randomized point-to-point route finding problems for which $h_{vw}$ and $c_{vw}$ were computed. In all experiments, the departure time (and thus the weather) was chosen randomly within a 3-day window. The departure and arrival coordinates were chosen randomly across Europe but within \unit[10--50]{km} of each other. Overall, the mean heuristic quality is only \unit[20]{\%}. Recall however that the heuristic is calculated as the straight line path under the best weather conditions in the whole time interval and area in which the flight happens. Often, there will be a tiny area with good weather in otherwise much worse weather such that $h_{vw}$ is small but $c_{vw}$ is large.

\Cref{fig:Results_PerformanceAnalysisHeuristic2} evaluates the influence of the heuristic quality $q_\text{heuristic}$ on the number of edge cost evaluations. The analysis consists of using the developed label correcting algorithm to solve 10 randomly generated problems for each combination of $q_\text{heuristic}$ and the number of areas of interest $N$. The resulting averages show that with increasing $N$ the heuristic becomes significantly more effective at reducing the required edge cost calculations. While the number of edge cost function evaluations is heavily dependent on the problem statement and the weather data, it is clear that especially large problems can be sped up if future research improves the heuristic. For example, for $N=7$ improving $q_\text{heuristic}$ to \unit[70]{\%} allows an order-of-magnitude reduction of edge cost computations. All in all, as shown in \cref{tab:Results_ComputationTimes}, MetPASS however already calculates station-keeping, point-to-point and large-scale multi-goal missions at sufficient grid resolution in below 20 minutes. The approach is thus sufficiently fast for mission analysis, pre-flight planning and in-flight re-planning. 

\section{Conclusion}
\label{sec:Conclusion}

This paper presented MetPASS, the Meteorology-aware Path Planning and Analysis Framework for Solar-powered UAVs. Using a combination of dynamic programming techniques and an A*-search-algorithm with a custom heuristic, optimal paths can be generated for large-scale station-keeping, point-to-point and multi-goal aerial inspection missions. In contrast to previous literature, MetPASS is the first framework which considers \emph{all} aspects that influence the safety and efficiency of solar-powered flight: By incorporating historical or forecasted meteorological data, MetPASS avoids environmental risks (thunderstorms, rain, humidity and icing, strong winds and wind gusts) and exploits advantageous regions (high sun radiation and tailwind). In addition, it avoids system risks such as low battery state of charge and returns safe paths through cluttered terrain. The loop to actual flight operations is closed by allowing direct waypoint upload to the aerial vehicle.

MetPASS has been applied for the feasibility analysis, pre-planning and re-planning of three different missions: \emph{AtlantikSolar}'s continuous \unit[2338]{km} and 81-hour world endurance record flight, a hypothetical \unit[4000]{km} Atlantic crossing from Newfoundland to Portugal, and a \unit[230]{km} two-glacier and \unit[580]{km} six-glacier remote sensing mission above the Arctic Ocean near Greenland. These missions have clearly shown that frameworks such as MetPASS, which combine an aircraft system model with meteorological data in a mathematically structured way, are indispensable for the reliable execution of large-scale solar-powered UAV missions. In addition, the following characteristics and lessons learned can be derived:
\begin{itemize}
\item \emph{Quality of weather data:} Only high quality weather data brings tangible benefits in mission planning. High confidence in the weather forecasts is thus required and the most up-to-date forecasts always have to be used. Wrong weather data can even lead to higher costs than the naive paths.
\item \emph{Cost advantages to naive solutions:} The cost advantages (i.e. safety and performance) generated by incorporating weather data decrease with
	\begin{itemize}
	\item Less weather variability, because less areas with above-average weather conditions exist and can be exploited by MetPASS. This especially applies to good macro weather situations.
	\item Smaller planning area, which also decreases weather variability.
	\item High or cluttered terrain, which limits the valid path choices.
	\end{itemize}
\end{itemize}

\bibliographystyle{plain}
\bibliography{refs/refs_all}   


\end{document}